\documentclass[11pt]{article}

\usepackage[a4paper,margin=0.75in]{geometry}
\usepackage[T1]{fontenc}
\usepackage[utf8]{inputenc}
\usepackage{lmodern}
\usepackage{authblk}
\usepackage{xcolor}
\usepackage{amsmath,amssymb,amsthm,mathtools,bm}
\usepackage{enumitem}
\usepackage{booktabs}
\usepackage{array}
\usepackage{longtable}
\usepackage{pdflscape}
\usepackage{graphicx}
\usepackage{tikz}
\usepackage{hyperref}
\usepackage{float}
\usepackage{multirow}
\usepackage{comment}

\hypersetup{
  colorlinks=true,
  hypertexnames=false,
  linkcolor=blue!50!black,
  citecolor=blue!50!black,
  urlcolor=blue
}

\newcommand{\dsC}{\mathbb{C}}

\newcommand{\dsH}{\mathbb{H}}

\newcommand{\dsN}{\mathbb{N}}

\newcommand{\dsR}{\mathbb{R}}

\newcommand{\scM}{\mathcal{M}}
\newcommand{\scN}{\mathcal{N}}

\newcommand{\scR}{\mathcal{R}}

\newcommand{\HomG}[2]{\operatorname{Hom}_G\!\left(#1,#2\right)}
\newcommand{\Hom}{\operatorname{Hom}}
\newcommand{\End}{\operatorname{End}}
\newcommand{\Sym}{\operatorname{Sym}}

\newcommand{\SO}{\operatorname{SO}}
\newcommand{\Orth}{\operatorname{O}}

\newcommand{\tr}{\operatorname{tr}}

\newcommand{\set}[1]{\{#1\}}

\usetikzlibrary{arrows.meta,backgrounds,calc,fit}

\makeatletter
\newif\ifiso@coreellipsis
\newif\ifiso@barecore
\newif\ifiso@leafenclosure
\tikzset{
  iso vertex/.style={circle,fill=black,inner sep=0pt,minimum size=3pt},
  iso wire/.style={draw=black,line width=.5pt},
  iso edge label/.style={midway,font=\small,inner sep=2pt,outer sep=0pt,
    fill=none,sloped=false},
  iso left edge label/.style={iso edge label,auto=left},
  iso right edge label/.style={iso edge label,auto=right},
  iso omission/.style={font=\small,inner sep=0pt,fill=none,sloped},
  iso input/.style={font=\small,inner sep=1pt,anchor=north,fill=none},
  iso metadata/.style={font=\footnotesize,inner sep=1pt,fill=none},
  iso group label/.style={font=\small,inner sep=1pt,anchor=south west,fill=none},
  iso bounds/.style={draw=none,fill=none,inner sep=0pt,outer sep=0pt},
  iso frame/.style={draw=black!35,densely dashed,rounded corners=2pt,
    fill=black!2,outer sep=0pt,inner sep=\iso@framepadding},
  iso leaf frame/.style={iso frame},
  iso core frame/.style={iso frame},
  iso cell/.style={draw=black,line width=.4pt,fill=none,inner sep=0pt,
    minimum width=\iso@tablesep cm,minimum height=\iso@tablesep cm,
    font=\iso@tableaufont},
  iso monochrome/.style={iso core frame/.append style={draw=black!60,fill=black!4}},
  iso slot sep/.store in=\iso@slotsep,
  iso output length/.store in=\iso@outputlength,
  iso core sep/.store in=\iso@coresep,
  iso tableau sep/.store in=\iso@tablesep,
  iso tableau rows/.store in=\iso@maxrows,
  iso tableau font/.store in=\iso@tableaufont,
  iso frame padding/.store in=\iso@framepadding,
  iso label gap/.store in=\iso@labelgap,
  iso tableau gap/.store in=\iso@tableaugap,
  iso core ellipses/.is if=iso@coreellipsis,
  iso diagram/.style={x=1cm,y=1cm,font=\small,
    baseline=(current bounding box.center),
    iso slot sep=.85,iso output length=.85,iso core sep=1.05,
    iso tableau sep=.34,iso tableau rows=0,iso tableau font=\footnotesize,
    iso frame padding=.5em,iso label gap=.5em,iso tableau gap=1em,
    iso core ellipses=false}
}
\pgfkeys{/iso leaf/.is family,/iso leaf,
  copies/.store in=\iso@copies,
  edge label/.store in=\iso@joinlabel,
  input omission/.store in=\iso@inputomission,
  tableau symbol/.store in=\iso@tableausymbol,
  enclosure/.is if=iso@leafenclosure,
  tableau label/.store in=\iso@tableaulabel}
\pgfkeys{/iso fuse/.is family,/iso fuse,
  ellipsis/.is choice,
  ellipsis/true/.code={\def\iso@edgeellipsis{1}},
  ellipsis/false/.code={\def\iso@edgeellipsis{0}},
  ellipsis/.default=true}
\newcount\iso@row
\newcount\iso@col
\newcount\iso@physical
\newcount\iso@degree
\newcount\iso@fusions
\newcount\iso@count
\newcount\iso@leafcount
\newsavebox{\iso@cellbox}
\newdimen\iso@measured

\newcommand{\iso@put}[2]{\expandafter\protected@edef\csname iso@#1\endcsname{#2}}
\newcommand{\iso@get}[1]{\csname iso@#1\endcsname}
\newcommand{\iso@error}[1]{\PackageError{tensor-diagrams}{#1}%
  {See docs/tensor-diagrams.md for the diagram syntax.}}
\newcommand{\iso@frame}[3]{%
  \begin{scope}[on background layer]
    \node[#3,fit/.expanded={#2}]
      (#1-frame) {};
  \end{scope}
}
\newcommand{\iso@edge}[5]{%
  \ifnum#5=1\relax
    \draw[iso wire] (#1) -- ($(#1)!.38!(#2)$);
    \path (#1) -- (#2) node[midway,iso omission] {$\cdots$};
    \draw[iso wire] ($(#1)!.62!(#2)$) -- (#2);
  \else
    \draw[iso wire] (#1) -- (#2);
  \fi
  \path (#1) -- (#2) node[iso #4 edge label]
    (iso-edge-#1-#2) {$#3$};
  \protected@edef\iso@networknodes{\iso@networknodes
    (#1)(#2)(iso-edge-#1-#2)}%
}
\NewDocumentEnvironment{isotypicdiagram}{O{} +b}{%
  \iso@drawdiagram{#1}{#2}%
}{}
\newcommand{\iso@drawdiagram}[2]{%
  \begin{tikzpicture}[iso diagram,#1]
  \iso@leafcount=0
  \def\iso@maxheight{0}%
  \def\iso@irreps{}%
  \def\iso@finished{0}%
  \def\iso@leafnodes{}%
  #2%
  \ifnum\iso@finished=0
    \iso@error{Missing \string\IsoCore\space or \string\IsoLeafOutput}%
  \fi
  \end{tikzpicture}%
}

\newcommand{\iso@youngbar}[5]{%
  \coordinate (#1-left) at ($(#3)+(0,#5)$);
  \coordinate (#1-right) at ($(#4)+(0,#5)$);
  \node[iso bounds,fit=(#1-left)(#1-right),inner xsep=2pt,
    minimum height=.12cm,iso wire,fill=#2] (#1) {};
  \protected@edef\iso@networknodes{\iso@networknodes (#1)}%
}
\newcommand{\iso@young}[2]{%
  \iso@count=1
  \@whilenum\iso@count<\numexpr\iso@degree+1\relax\do{%
    \edef\iso@slotname{s\the\iso@count}%
    \coordinate (#1-in-\iso@slotname) at
      ({#2+(\the\iso@count-(\the\iso@degree+1)/2)*\iso@slotsep},-.94);
    \draw[iso wire] (#1-in-\iso@slotname)
      -- ([yshift=.28cm]#1-in-\iso@slotname)
      -- ([yshift=-.28cm]#1-\iso@slotname) -- (#1-\iso@slotname);
    \protected@edef\iso@networknodes{\iso@networknodes (#1-in-\iso@slotname)}%
    \advance\iso@count by1
  }%
  \iso@row=1
  \@whilenum\iso@row<\numexpr\iso@rows+1\relax\do{%
    \edef\iso@first{s\iso@get{#1-box-\the\iso@row-1}}%
    \edef\iso@last{s\iso@get{#1-box-\the\iso@row-\iso@get{#1-row-width-\the\iso@row}}}%
    \iso@youngbar{#1-row-\the\iso@row}{white}
      {#1-in-\iso@first}{#1-in-\iso@last}{.18}%
    \advance\iso@row by1
  }%
  \iso@col=1
  \@whilenum\iso@col<\numexpr\iso@width+1\relax\do{%
    \edef\iso@first{s\iso@get{#1-box-1-\the\iso@col}}%
    \edef\iso@last{\iso@get{#1-column-last-\the\iso@col}}%
    \iso@youngbar{#1-column-\the\iso@col}{black}
      {#1-\iso@first}{#1-\iso@last}{-.18}%
    \advance\iso@col by1
  }%
}
\newcommand{\IsoLeaf}[6][]{%
  \ifnum\iso@finished=1 \iso@error{Leaves must precede the core}\fi
  \ifcsname iso@leaf-#2\endcsname \iso@error{Duplicate leaf name #2}\fi
  \def\iso@candidate{#4}%
  \@for\iso@prior:=\iso@irreps\do{%
    \ifx\iso@candidate\iso@prior
      \iso@error{Use one leaf per active isotypic factor}%
    \fi
  }%
  \ifx\iso@irreps\@empty
    \def\iso@irreps{{#4}}%
  \else
    \protected@edef\iso@irreps{\iso@irreps,{#4}}%
  \fi
  \iso@put{leaf-#2}{1}%
  \advance\iso@leafcount by1
  \def\iso@prefix{#2}%
  \coordinate (#2-centre) at (#3,0);
  \pgfkeys{/iso leaf/.cd,copies={},edge label={},input omission={},tableau symbol={},
    enclosure=true,tableau label={T_{#4}=},#1}%
  \iso@put{#2-tableau-symbol}{\iso@tableausymbol}%
  \iso@put{#2-tableau-label}{\iso@tableaulabel}%
  \iso@count=0
  \@for\iso@copy:=\iso@copies\do{%
    \advance\iso@count by1
    \iso@put{#2-copy-\the\iso@count}{\iso@copy}%
  }%
  \iso@row=0 \iso@degree=0
  \def\iso@tablerows{#5}%
  \def\iso@previouswidth{2147483647}%
  \@for\iso@entries:=\iso@tablerows\do{%
    \advance\iso@row by1 \iso@col=0
    \def\iso@previousentry{0}%
    \@for\iso@entry:=\iso@entries\do{%
      \advance\iso@col by1 \advance\iso@degree by1
      \ifnum\iso@entry<1 \iso@error{Tableau entries must be positive integers}\fi
      \ifnum\iso@entry<\iso@previousentry
        \iso@error{Tableau rows must weakly increase}%
      \fi
      \edef\iso@previousentry{\iso@entry}%
      \ifnum\iso@row>1
        \edef\iso@above{#2-box-\the\numexpr\iso@row-1\relax-\the\iso@col}%
        \ifcsname iso@\iso@above\endcsname
          \ifnum\iso@entry>\iso@get{#2-entry-\iso@get{\iso@above}}\relax\else
            \iso@error{Tableau columns must strictly increase}%
          \fi
        \fi
      \fi
      \iso@put{#2-box-\the\iso@row-\the\iso@col}{\the\iso@degree}%
      \iso@put{#2-entry-\the\iso@degree}{\iso@entry}%
    }%
    \ifnum\iso@col=0 \iso@error{Tableau rows must be nonempty}\fi
    \iso@put{#2-row-width-\the\iso@row}{\the\iso@col}%
    \ifnum\iso@col>\iso@previouswidth
      \iso@error{Tableau row lengths must weakly decrease}%
    \fi
    \edef\iso@previouswidth{\the\iso@col}%
    \ifnum\iso@row=1
      \edef\iso@width{\the\iso@col}%
    \fi
  }%
  \ifnum\iso@degree=0
    \iso@error{Empty leaf #2: omit degree-zero isotypic factors}%
  \fi
  \edef\iso@rows{\the\iso@row}%
  \def\iso@networknodes{}%
  \iso@physical=0 \iso@col=1
  \@whilenum\iso@col<\numexpr\iso@width+1\relax\do{%
    \iso@row=1
    \@whilenum\iso@row<\numexpr\iso@rows+1\relax\do{%
      \ifcsname iso@#2-box-\the\iso@row-\the\iso@col\endcsname
        \advance\iso@physical by1
        \edef\iso@slotname{s\iso@get{#2-box-\the\iso@row-\the\iso@col}}%
        \iso@put{#2-column-last-\the\iso@col}{\iso@slotname}%
        \edef\iso@entry{\iso@get{#2-entry-\iso@get{#2-box-\the\iso@row-\the\iso@col}}}%
        \pgfmathsetmacro\iso@x{#3+(\the\iso@physical-(\the\iso@degree+1)/2)*\iso@slotsep}%
        \iso@put{#2-\iso@slotname-x}{\iso@x}%
        \iso@put{#2-\iso@slotname-y}{0}%
        \iso@put{#2-\iso@slotname-spin}{#4}%
        \iso@put{#2-\iso@slotname-min}{\iso@x}%
        \iso@put{#2-\iso@slotname-max}{\iso@x}%
        \coordinate (#2-\iso@slotname) at (\iso@x,0);
        \protected@edef\iso@networknodes{\iso@networknodes
          (#2-\iso@slotname)}%
        \edef\iso@root{\iso@slotname}%
      \fi
      \advance\iso@row by1
    }%
    \advance\iso@col by1
  }%
  \iso@fusions=0
  #6%
  \ifnum\iso@fusions=\numexpr\iso@degree-1\relax\else
    \iso@error{Leaf #2 must have exactly d-1 fusions}%
  \fi
  \iso@put{#2-root}{\iso@root}%
  \iso@put{#2-lambda}{#4}%
  \iso@put{#2-mu}{\iso@get{#2-\iso@root-spin}}%
  \ifx\iso@joinlabel\@empty
    \iso@put{#2-edge-label}{\iso@get{#2-mu}}%
  \else
    \iso@put{#2-edge-label}{\iso@joinlabel}%
  \fi
  \iso@put{#2-rootx}{\iso@get{#2-\iso@root-x}}%
  \iso@put{#2-rooty}{\iso@get{#2-\iso@root-y}}%
  \coordinate (#2-out) at (#2-\iso@root);
  \iso@young{#2}{#3}%
  \def\iso@leafinputs{}%
  \iso@physical=1
  \@whilenum\iso@physical<\numexpr\iso@degree+1\relax\do{%
    \edef\iso@slotname{s\the\iso@physical}%
    \edef\iso@entry{\iso@get{#2-entry-\the\iso@physical}}%
    \ifcsname iso@#2-copy-\iso@entry\endcsname
      \node[iso input] (#2-input-\iso@slotname)
        at ([yshift=-\iso@labelgap]#2-in-\iso@slotname)
        {$\iso@get{#2-copy-\iso@entry}$};
    \else
      \node[iso input] (#2-input-\iso@slotname)
        at ([yshift=-\iso@labelgap]#2-in-\iso@slotname) {$x_{#4,\iso@entry}$};
    \fi
    \protected@edef\iso@leafinputs{\iso@leafinputs (#2-input-\iso@slotname)}%
    \advance\iso@physical by1
  }%
  \ifx\iso@inputomission\@empty\else
    \coordinate (#2-input-centre) at (#2-out |- #2-in-s1);
    \node[iso input] (#2-input-omission)
      at ([yshift=-\iso@labelgap]#2-input-centre) {$\iso@inputomission$};
    \protected@edef\iso@leafinputs{\iso@leafinputs (#2-input-omission)}%
  \fi
  \node[iso bounds,fit/.expanded={\iso@leafinputs}] (#2-inputs) {};
  \node[iso bounds,fit/.expanded={\iso@networknodes (#2-inputs)}]
    (#2-labelled-tree) {};
  \ifiso@leafenclosure
    \iso@frame{#2}{(#2-labelled-tree)}{iso leaf frame}%
    \node[iso group label] (#2-label)
      at ([yshift=\iso@labelgap]#2-frame.north west) {$L_{#4}$};
    \pgfextracty{\iso@measured}{\pgfpointanchor{#2-label}{north}}%
  \else
    \node[iso bounds,fit=(#2-labelled-tree)] (#2-frame) {};
    \pgfextracty{\iso@measured}{\pgfpointanchor{#2-frame}{north}}%
  \fi
  \pgfmathsetmacro\iso@maxheight{max(\iso@maxheight,\the\iso@measured/1cm)}%
  \protected@edef\iso@leafnodes{\iso@leafnodes (#2-frame)}%
  \iso@put{#2-centre}{#3}%
  \iso@put{#2-rows}{\iso@rows}%
  \iso@put{#2-width}{\iso@width}%
  \iso@put{#2-tablerows}{\iso@tablerows}%
  \pgfmathsetmacro\iso@maxrows{max(\iso@maxrows,\iso@rows)}%
}

\newcommand{\iso@measuretableau}[1]{%
  \protected@edef\iso@tablerows{\iso@get{#1-tablerows}}%
  \@for\iso@entries:=\iso@tablerows\do{%
    \@for\iso@entry:=\iso@entries\do{%
      \sbox{\iso@cellbox}{\pgfinterruptpicture
        \normalfont\iso@tableaufont$\iso@entry$\endpgfinterruptpicture}%
      \pgfmathsetmacro\iso@tablesep{max(\iso@tablesep,
        (\the\wd\iso@cellbox+\iso@labelgap)/1cm,
        (\the\ht\iso@cellbox+\the\dp\iso@cellbox+\iso@labelgap)/1cm)}%
    }%
  }%
}

\newcommand{\iso@metadata}[1]{%
  \edef\iso@rows{\iso@get{#1-rows}}%
  \edef\iso@width{\iso@get{#1-width}}%
  \protected@edef\iso@tablerows{\iso@get{#1-tablerows}}%
  \coordinate (#1-tableau-centre) at (#1-centre |- iso-tableau-mid);
  \protected@edef\iso@tableausymbol{\iso@get{#1-tableau-symbol}}%
  \ifx\iso@tableausymbol\@empty
  \iso@row=0
  \@for\iso@entries:=\iso@tablerows\do{%
    \advance\iso@row by1 \iso@col=0
    \@for\iso@entry:=\iso@entries\do{%
      \advance\iso@col by1
      \node[iso cell] (#1-tableau-\the\iso@row-\the\iso@col) at
        ($(#1-tableau-centre)+
         ({(\the\iso@col-(\iso@width+1)/2)*\iso@tablesep},
          {((\iso@rows+1)/2-\the\iso@row)*\iso@tablesep})$) {$\iso@entry$};
    }%
  }%
  \coordinate (#1-tableau-west) at
    ($(#1-tableau-centre)+({-\iso@width*\iso@tablesep/2},0)$);
  \node[iso metadata,anchor=east] at ([xshift=-\iso@labelgap]#1-tableau-west)
    {$\iso@get{#1-tableau-label}$};
  \else
    \node[iso metadata] at (#1-tableau-centre)
      {$\iso@tableausymbol$};
  \fi
}

\newcommand{\IsoFuse}[5][]{%
  \ifnum\iso@finished=1 \iso@error{Fusions must precede the output}\fi
  \pgfkeys{/iso fuse/.cd,ellipsis=false,#1}%
  \ifcsname iso@\iso@prefix-#2-spin\endcsname
    \iso@error{Duplicate node #2 in leaf \iso@prefix}%
  \fi
  \@for\iso@child:=#3,#4\do{%
    \ifcsname iso@\iso@prefix-\iso@child-spin\endcsname\else
      \iso@error{Unknown child \iso@child\space in leaf \iso@prefix}%
    \fi
    \ifcsname iso@\iso@prefix-\iso@child-used\endcsname
      \iso@error{Child \iso@child\space has already been coupled}%
    \fi
    \iso@put{\iso@prefix-\iso@child-used}{1}%
  }%
  \ifdim\iso@get{\iso@prefix-#3-max}pt<\iso@get{\iso@prefix-#4-min}pt\else
    \iso@error{Couple adjacent subtrees in left-to-right order}%
  \fi
  \pgfmathsetmacro\iso@xmin{min(\iso@get{\iso@prefix-#3-min},\iso@get{\iso@prefix-#4-min})}%
  \pgfmathsetmacro\iso@xmax{max(\iso@get{\iso@prefix-#3-max},\iso@get{\iso@prefix-#4-max})}%
  \pgfmathsetmacro\iso@x{(\iso@xmin+\iso@xmax)/2}%
  \pgfmathsetmacro\iso@y{(\iso@xmax-\iso@xmin)/2}%
  \iso@put{\iso@prefix-#2-x}{\iso@x}%
  \iso@put{\iso@prefix-#2-y}{\iso@y}%
  \iso@put{\iso@prefix-#2-min}{\iso@xmin}%
  \iso@put{\iso@prefix-#2-max}{\iso@xmax}%
  \iso@put{\iso@prefix-#2-spin}{#5}%
  \node[iso vertex] (\iso@prefix-#2) at (\iso@x,\iso@y) {};
  \iso@edge{\iso@prefix-#3}{\iso@prefix-#2}
    {\iso@get{\iso@prefix-#3-spin}}{left}{\iso@edgeellipsis}%
  \iso@edge{\iso@prefix-#4}{\iso@prefix-#2}
    {\iso@get{\iso@prefix-#4-spin}}{right}{\iso@edgeellipsis}%
  \def\iso@root{#2}%
  \advance\iso@fusions by1
}

\newenvironment{tensordiagram}[1][]{%
  \begin{tikzpicture}[iso diagram,#1]
  \def\iso@prefix{tensor}%
  \def\iso@networknodes{}%
  \def\iso@finished{0}%
  \iso@degree=0 \iso@fusions=0
}{%
  \ifnum\iso@finished=0
    \iso@error{Missing \string\TensorOutput\space or \string\TensorTrain}%
  \fi
  \end{tikzpicture}%
}
\newcommand{\TensorInput}[4]{%
  \ifnum\iso@finished=1 \iso@error{Inputs must precede the output}\fi
  \ifnum\iso@fusions>0 \iso@error{Inputs must precede all fusions}\fi
  \ifcsname iso@tensor-#1-spin\endcsname
    \iso@error{Duplicate input #1}%
  \fi
  \pgfmathsetmacro\iso@x{#2*\iso@slotsep}%
  \iso@put{tensor-#1-x}{\iso@x}%
  \iso@put{tensor-#1-y}{0}%
  \iso@put{tensor-#1-min}{\iso@x}%
  \iso@put{tensor-#1-max}{\iso@x}%
  \iso@put{tensor-#1-spin}{#4}%
  \coordinate (tensor-#1) at (\iso@x,0);
  \node[iso input] at ([yshift=-\iso@labelgap]tensor-#1) {$#3$};
  \def\iso@root{#1}%
  \advance\iso@degree by1
}
\newcommand{\TensorOutput}[1]{%
  \ifnum\iso@finished=1 \iso@error{Only one output is allowed}\fi
  \ifnum\iso@fusions=\numexpr\iso@degree-1\relax\else
    \iso@error{A tensor tree must have exactly d-1 fusions}%
  \fi
  \def\iso@finished{1}%
  \coordinate (tensor-output) at
    ($(tensor-\iso@root)+(0,\iso@outputlength)$);
  \draw[iso wire,-{Stealth[length=3pt,width=3pt]}]
    (tensor-\iso@root) -- (tensor-output)
    node[iso right edge label] {$\iso@get{tensor-\iso@root-spin}$};
  \node[iso metadata,anchor=south,font=\small]
    at ([yshift=\iso@labelgap]tensor-output) {$#1$};
}

\newcommand{\TensorTrain}[2]{%
  \ifnum\iso@fusions>0
    \iso@error{A standalone train supplies its own fusions}%
  \fi
  \iso@barecoretrue
  \iso@leafcount=\iso@degree
  \def\iso@maxheight{0}%
  \@for\iso@input:=#1\do{%
    \ifcsname iso@tensor-\iso@input-spin\endcsname\else
      \iso@error{Unknown train input \iso@input}%
    \fi
    \iso@put{leaf-\iso@input}{1}%
    \iso@put{\iso@input-rootx}{\iso@get{tensor-\iso@input-x}}%
    \iso@put{\iso@input-mu}{\iso@get{tensor-\iso@input-spin}}%
    \coordinate (\iso@input-out) at (tensor-\iso@input);
  }%
  \IsoCore{#1}{#2}%
  \iso@barecorefalse
}
\newcommand{\IsoTrain}[3]{%
  \ifnum\iso@finished=1 \iso@error{Trains must precede the output}\fi
  \ifcsname iso@\iso@prefix-#1-spin\endcsname
    \iso@error{Duplicate node #1 in leaf \iso@prefix}%
  \fi
  \def\iso@traincount{0}%
  \def\iso@trainheight{0}%
  \@for\iso@child:=#2\do{%
    \ifcsname iso@\iso@prefix-\iso@child-spin\endcsname\else
      \iso@error{Unknown train input \iso@child}%
    \fi
    \ifcsname iso@\iso@prefix-\iso@child-used\endcsname
      \iso@error{Child \iso@child\space has already been coupled}%
    \fi
    \iso@put{\iso@prefix-\iso@child-used}{1}%
    \ifnum\iso@traincount=0
      \edef\iso@trainmin{\iso@get{\iso@prefix-\iso@child-min}}%
    \else
      \ifdim\iso@trainmax pt<\iso@get{\iso@prefix-\iso@child-min}pt\else
        \iso@error{Train inputs must be disjoint and in column order}%
      \fi
    \fi
    \edef\iso@trainmax{\iso@get{\iso@prefix-\iso@child-max}}%
    \edef\iso@trainlastx{\iso@get{\iso@prefix-\iso@child-x}}%
    \pgfmathsetmacro\iso@trainheight{max(\iso@trainheight,
      \iso@get{\iso@prefix-\iso@child-y})}%
    \edef\iso@traincount{\the\numexpr\iso@traincount+1\relax}%
  }%
  \ifnum\iso@traincount=0 \iso@error{A train needs at least one input}\fi
  \@for\iso@spin:=#3\do{\protected@edef\iso@trainspin{\iso@spin}}%
  \pgfmathsetmacro\iso@trainoutx{\iso@trainlastx+\iso@outputlength}%
  \pgfmathsetmacro\iso@trainouty{\iso@trainheight+\iso@coresep}%
  \begin{scope}
    \iso@barecoretrue
    \iso@leafcount=\iso@traincount\relax
    \edef\iso@maxheight{\iso@trainheight}%
    \@for\iso@child:=#2\do{%
      \iso@put{leaf-\iso@child}{1}%
      \iso@put{\iso@child-rootx}{\iso@get{\iso@prefix-\iso@child-x}}%
      \iso@put{\iso@child-mu}{\iso@get{\iso@prefix-\iso@child-spin}}%
      \coordinate (\iso@child-out) at (\iso@prefix-\iso@child);
    }%
    \IsoCore{#2}{#3}%
    \coordinate (\iso@prefix-#1) at (iso-output);
    \node[iso bounds,
      fit/.expanded={\iso@networknodes (iso-output)(iso-output-label)}]
      (\iso@prefix-#1-train) {};
  \end{scope}
  \iso@put{\iso@prefix-#1-x}{\iso@trainoutx}%
  \iso@put{\iso@prefix-#1-y}{\iso@trainouty}%
  \iso@put{\iso@prefix-#1-min}{\iso@trainmin}%
  \iso@put{\iso@prefix-#1-max}{\iso@trainmax}%
  \iso@put{\iso@prefix-#1-spin}{\iso@trainspin}%
  \iso@put{\iso@prefix-#1-hasoutput}{1}%
  \protected@edef\iso@networknodes{\iso@networknodes (\iso@prefix-#1-train)}%
  \def\iso@root{#1}%
  \advance\iso@fusions by\numexpr\iso@traincount-1\relax
}
\newcommand{\IsoLeafOutput}[1]{%
  \ifnum\iso@finished=1 \iso@error{Only one output is allowed}\fi
  \ifnum\iso@leafcount=1\else
    \iso@error{A standalone Schur diagram must contain exactly one leaf}%
  \fi
  \edef\iso@leafroot{\iso@get{#1-root}}%
  \ifcsname iso@#1-\iso@leafroot-hasoutput\endcsname\else
    \coordinate (#1-output) at ($(#1-out)+(0,\iso@outputlength)$);
    \draw[iso wire,-{Stealth[length=3pt,width=3pt]}]
      (#1-out) -- (#1-output)
      node[iso right edge label] {$\iso@get{#1-mu}$};
  \fi
  \def\iso@finished{1}%
  \iso@measuretableau{#1}%
  \coordinate (iso-tableau-mid) at
    ($(#1-frame.south)+(0,-\iso@tableaugap)
      +(0,{-\iso@maxrows*\iso@tablesep/2})$);
  \iso@metadata{#1}%
}
\newcommand{\IsoCore}[3][]{%
  \ifnum\iso@leafcount=0
    \iso@error{Use the scalar 1 for the empty diagram}%
  \fi
  \ifnum\iso@finished=1 \iso@error{Only one core is allowed per diagram}\fi
  \def\iso@finished{1}%
  \iso@count=0
  \@for\iso@spin:=#3\do{%
    \advance\iso@count by1
    \iso@put{core-spin-\the\iso@count}{\iso@spin}%
  }%
  \edef\iso@pathlength{\the\iso@count}%
  \pgfmathsetmacro\iso@corey{\iso@maxheight+\iso@coresep}%
  \def\iso@networknodes{}%
  \iso@count=0
  \@for\iso@leaf:=#2\do{%
    \advance\iso@count by1
    \ifcsname iso@leaf-\iso@leaf\endcsname\else
      \iso@error{Unknown core input \iso@leaf}%
    \fi
    \ifcsname iso@\iso@leaf-core-used\endcsname
      \iso@error{Duplicate core input \iso@leaf}%
    \fi
    \iso@put{\iso@leaf-core-used}{1}%
    \edef\iso@x{\iso@get{\iso@leaf-rootx}}%
    \ifnum\iso@count=1
      \edef\iso@firstx{\iso@x}%
      \coordinate (iso-core-1) at (\iso@x,\iso@corey);
      \edef\iso@firstmu{\iso@get{\iso@leaf-mu}}%
    \else
      \ifdim\iso@x pt>\iso@previousx pt\else
        \iso@error{Core inputs must be placed from left to right}%
      \fi
      \node[iso vertex] (iso-core-\the\iso@count) at (\iso@x,\iso@corey) {};
      \edef\iso@previous{\the\numexpr\iso@count-1\relax}%
      \ifnum\iso@count=2
        \iso@edge{iso-core-1}{iso-core-2}{}{left}
          {\ifiso@coreellipsis1\else0\fi}%
      \else
        \iso@edge{iso-core-\iso@previous}{iso-core-\the\iso@count}
          {\iso@get{core-spin-\the\numexpr\iso@count-2\relax}}{left}
          {\ifiso@coreellipsis1\else0\fi}%
      \fi
    \fi
    \ifiso@barecore
      \iso@edge{\iso@leaf-out}{iso-core-\the\iso@count}
        {\iso@get{\iso@leaf-mu}}{right}{0}%
    \else
      \draw[iso wire] (\iso@leaf-out) -- (iso-core-\the\iso@count);
    \fi
    \edef\iso@previousx{\iso@x}%
  }%
  \ifnum\iso@count=\iso@leafcount\else
    \iso@error{The core must use every leaf exactly once}%
  \fi
  \ifnum\iso@count=1
    \def\iso@expected{1}%
    \edef\iso@finalmu{\iso@get{core-spin-1}}%
    \ifx\iso@firstmu\iso@finalmu\else
      \iso@error{An identity core cannot change its irreducible representation label}%
    \fi
  \else
    \edef\iso@expected{\the\numexpr\iso@count-1\relax}%
  \fi
  \ifnum\iso@pathlength=\iso@expected\else
    \iso@error{Incorrect number of core output labels}%
  \fi
  \coordinate (iso-output) at ({\iso@x+\iso@outputlength},\iso@corey);
  \draw[iso wire,-{Stealth[length=3pt,width=3pt]}]
    (iso-core-\the\iso@count) -- (iso-output)
    node[iso left edge label] (iso-output-label)
      {$\iso@get{core-spin-\iso@pathlength}$};
  \ifiso@barecore\else
  \node[iso bounds,
    fit/.expanded={\iso@networknodes (iso-core-1)(iso-output)(iso-output-label)}]
    (iso-core-content) {};
  \path let \p1=(iso-core-content.south west),
    \p2=(iso-core-content.north east), \p3=(iso-core-1),
    \n1={max(\y3-\y1,\y2-\y3)} in
    coordinate (iso-core-sw) at ({\x1-\n1},{\y3-\n1})
    coordinate (iso-core-ne) at ({\x2+\n1},{\y3+\n1});
  \iso@frame{iso-core}{(iso-core-sw)(iso-core-ne)}{iso core frame}%
  \node[iso metadata,anchor=west]
    at ([xshift=\iso@labelgap]iso-core-frame.east) {$#1$};
  \node[iso group label,anchor=south] (iso-core-label)
    at ([yshift=\iso@labelgap]iso-core-frame.north)
    {$C$};
  \@for\iso@leaf:=#2\do{%
    \coordinate (iso-join-bottom) at (\iso@leaf-frame.north -| \iso@leaf-out);
    \coordinate (iso-join-top) at (iso-core-frame.south -| \iso@leaf-out);
    \path (iso-join-bottom) -- (iso-join-top)
      node[iso right edge label] {$\iso@get{\iso@leaf-edge-label}$};
    \iso@measuretableau{\iso@leaf}%
  }%
  \node[iso bounds,fit/.expanded={\iso@leafnodes}] (iso-leaves) {};
  \coordinate (iso-tableau-mid) at
    ($(iso-leaves.south)+(0,-\iso@tableaugap)
      +(0,{-\iso@maxrows*\iso@tablesep/2})$);
  \@for\iso@leaf:=#2\do{\iso@metadata{\iso@leaf}}%
  \fi
}
\newcommand{\IsoSchematicLeaf}[5][]{%
  \IsoLeaf[copies={{p_{#4,a_1}(x)},{},{},{p_{#4,a_{d_{#4}}}(x)}},
    input omission={\cdots},
    #1]{#2}{#3}{#4}{{1,2},{3,4}}{%
    \IsoFuse{left}{s1}{s3}{}%
    \IsoFuse{right}{s2}{s4}{}%
    \IsoFuse[ellipsis]{root}{left}{right}{#5}%
  }%
  \node[iso metadata] at ({#3},-.76) {$\cdots$};
  \node[iso metadata] at ({#3},-.18) {$\cdots$};
  \iso@put{#2-rows}{3}%
  \iso@put{#2-width}{4}%
  \iso@put{#2-tablerows}{%
    {a_1,a_2,a_3,\cdots},%
    {\vdots,\ddots,\vdots},%
    {\cdots,a_{d_{#4}}}}%
  \pgfmathsetmacro\iso@tablesep{max(\iso@tablesep,.70)}%
  \pgfmathsetmacro\iso@maxrows{max(\iso@maxrows,3)}%
}

\newcommand{\IsoLeafGap}[2]{%
  \node[iso metadata] at ($(#1-frame.east)!.5!(#2-frame.west)$)
    {$\cdots$};
}

\newcommand{\IsoAbstractDiagram}{%
  \begin{isotypicdiagram}[iso core ellipses,
    iso input/.append style={font=\scriptsize},
    iso edge label/.append style={font=\footnotesize}]
    \IsoSchematicLeaf{first}{0}{\alpha}{\mu_\alpha}
    \node[iso bounds,fit=(first-frame)(first-inputs)(first-label)]
      (iso-leaf-template) {};
    \pgfextractx{\iso@measured}{\pgfpointdiff
      {\pgfpointanchor{iso-leaf-template}{west}}
      {\pgfpointanchor{iso-leaf-template}{east}}}%
    \sbox{\iso@cellbox}{\pgfinterruptpicture
      \normalfont\footnotesize$\cdots$\endpgfinterruptpicture}%
    \pgfmathsetmacro\iso@leafstep{
      (\the\iso@measured+\the\wd\iso@cellbox+2*\iso@labelgap)/1cm}%
    \IsoSchematicLeaf{middle}{\iso@leafstep}{\lambda}{\mu_\lambda}
    \IsoSchematicLeaf{last}{2*\iso@leafstep}{\omega}{\mu_\omega}
    \IsoLeafGap{first}{middle}
    \IsoLeafGap{middle}{last}
    \IsoCore{first,middle,last}{{},\nu}
    \node[anchor=south] at ([yshift=\iso@labelgap]iso-core-label.north)
      {$Q(x^{\otimes D})=C\!\left(\bigotimes_\lambda
        L_\lambda(T_\lambda(x))\right)$};
  \end{isotypicdiagram}%
}
\makeatother

\newtheorem{theorem}{Theorem}[section]
\newtheorem{proposition}[theorem]{Proposition}
\newtheorem{lemma}[theorem]{Lemma}
\newtheorem{corollary}[theorem]{Corollary}
\newtheorem{algorithm}[theorem]{Algorithm}
\newtheorem{assumption}{Assumption}[section]

\theoremstyle{definition}
\newtheorem{definition}[theorem]{Definition}
\newtheorem{example}[theorem]{Example}

\title{Equivariant Modeling with Polynomial Features \\ via Minimal Homogeneous Generators}
\author[1]{Michael Liaofan Liu\thanks{Email: \texttt{michael.liu.2024@utexas.edu}}}
\author[1,2]{Joe Kileel}
\affil[1]{Oden Institute, University of Texas at Austin}
\affil[2]{Department of Mathematics, University of Texas at Austin}
\date{}

\begin{document}
\maketitle

\begin{abstract}

  We present a general framework for modeling and learning equivariant functions between finite-dimensional semisimple real group representations, unifying constructions developed across many application areas.
  Our model is a linear combination of equivariant polynomial features whose coefficients are arbitrary learnable functions of invariant polynomial features.
  We take as features minimal homogeneous generators for the invariant algebra and the covariant module over the invariant algebra, up to prescribed degrees.
  We provide a  \href{https://github.com/MichaelLiu2024/Equivariant-Polynomial-Tensor-Functions}{Python package} 
  that automatically computes these features and fits the resulting models to training data.
 The algorithm for feature computation requires explicit isotypic decompositions of the input and output representations, along with explicit Clebsch-Gordan decompositions of the tensor product of irreducible representations, and achieves speedups over alternative methods in computational algebra. %
Numerical experiments in scientific modeling demonstrate the full learning pipeline and yield accurate predictors with few trainable parameters, using the model both for end-to-end prediction and within layers of larger architectures.

\end{abstract}

\section{Introduction}\label{sec:intro}

Many physical systems obey natural symmetries which learned models should respect. For example, changing coordinates in a system of charged particles should affect
predicted accelerations according to the change of coordinates, while leaving predicted
energies unchanged. 
Such transformation laws appear throughout science and engineering, and should constrain the class of admissible models.

Equivariance formalizes this requirement. Let $G$ be a group acting on an input space
$X$ and an output space $Y$. In data-driven applications we seek to learn a function $F:X\to Y$ that is
\emph{$G$-equivariant}, meaning that
\[
  F(g \cdot x)=g \cdot F(x)
\]
for all $g\in G$ and $x\in X$, i.e., transforming the input and
then predicting gives the same result as predicting and then
transforming the output.
When the action on $Y$ leaves every output unchanged, this condition
reduces to \emph{$G$-invariance}.

A wide variety of approaches exist to encourage or enforce equivariance in machine learning models.
For instance, data augmentation \cite{SimardSteinkrausPlatt2003,ChenDobribanLee2020} encourages equivariance by training the model on many transformed copies of
input-output pairs, but it does not guarantee equivariance in general. Equivariant multilayer perceptrons
\cite{FinziWellingWilson2021} enforce equivariance by solving linear
constraints derived from group generators to parameterize equivariant linear
layers. %
Equivariant graph networks
\cite{ThomasEtAl2018,SatorrasHoogeboomWelling2021} enforce equivariance through
message passing. %

We investigate an alternative approach that enforces equivariance through polynomial features \cite{LingKurzawskiTempleton2016,Yarotsky2022,VillarEtAl2021}.
The underlying model is a linear combination of equivariant polynomials, called
\emph{covariants}, where the scalar coefficients are arbitrary learned functions of
invariant polynomials, called \emph{invariants}.

To formalize this approach, suppose that $X$ and $Y$ are finite-dimensional semisimple
real $G$-representations \cite{FH1991}.
Suppose first that the target function $F$ is a polynomial.
Then $F$ lies in the \emph{covariant module}
\[
  \scM(X,Y)\coloneq
  \{\text{$G$-equivariant polynomial maps from }X\text{ to }Y\}.
\]
This is a module over the \emph{invariant algebra}
\[
  \scR(X)\coloneq
  \{\text{$G$-invariant polynomials from }X\text{ to }\dsR\},
\]
which is an algebra over $\dsR$.
By Hilbert's finiteness theorem (see Lemmas~\ref{lem:invariant-finite-generation}
and \ref{lem:covariant-finite-generation}), there exist finite generating sets
$\set{r_1,\dots,r_R}$ for $\scR(X)$ as an $\dsR$-algebra and
$\set{m_1,\dots,m_M}$ for $\scM(X,Y)$ as an $\scR(X)$-module.
Thus, there exists a
polynomial map $p\colon\dsR^R\to\dsR^M$ such that
\begin{equation}\label{eq:polynomial-ansatz}
  F(x)=\sum_{i=1}^M p_i(r_1(x),\dots,r_R(x))m_i(x).
\end{equation}
For arbitrary targets $F$, we propose to replace the polynomial $p$ with a learnable function
$\scN_\theta\colon\dsR^R\to\dsR^M$, such as a neural network,
a random forest, or a kernel regression model. This yields a class of $G$-equivariant functions
\begin{equation}\label{eq:neural-ansatz}
  F_\theta(x)=\sum_{i=1}^M
  \scN_{\theta,i}(r_1(x),\dots,r_R(x))m_i(x),
\end{equation}
where the goal is to learn $\theta$ so that $F_\theta \approx F$.
The ansatz can serve as an end-to-end predictor or as an equivariant component within a larger architecture.
This model class has appeared before in various forms often under the name of ``scalarization'' \cite{Yarotsky2022, VillarEtAl2021, BlumSmithVillar2023, ZhangEtAl2026,KommalapatiEtAl2026}, but to our knowledge it has not yet been developed at  our level of generality and packaged into a user-friendly toolkit.  
The work~\cite{KommalapatiEtAl2026} precedes this one and is most similar.  
After introducing Eq.~\eqref{eq:neural-ansatz}, it  focused on the case of tensor-valued inputs/outputs,  and developed algorithms using binary forms for important applications to fluid dynamics.

We observe that the models $F_\theta$ are automatically equivariant, because the maps $m_i$ are
equivariant and polynomials $r_j$ are invariant. 
Consequently, the learnable coefficient functions $\scN_\theta$ need not satisfy any additional equivariance constraints and can be chosen at will.
Thus, the number of parameters in $\scN_\theta$ can be much smaller than in other equivariant networks.
Moreover, for fixed $G$, $X$, and $Y$, the same polynomial features may be reused to learn several target functions $F_1, F_2, \ldots$ by
training separate coefficient models $\scN_{\theta_1}, \scN_{\theta_2}, \ldots$ on different datasets. Finally, for polynomial targets Eq.~\eqref{eq:polynomial-ansatz} is linear in its parameters and can be fitted by linear least squares, while   Eq.~\eqref{eq:neural-ansatz} can be more interpretable than alternatives in machine learning if its coefficient functions are interpretable.

The next proposition characterizes the approximation power of
Eq.~\eqref{eq:neural-ansatz}. %

\begin{proposition}
\label{prop:degree-truncation}
Let $r_1,\ldots,r_R \in \mathcal{R}(X)$ and
$m_1,\ldots,m_M\in \mathcal{M}(X,Y)$.
Assume that the coefficient models
$\mathcal{N}_\theta:\mathbb{R}^R\to\mathbb{R}^M$
can approximate every continuous map uniformly on compact sets.
Let $F:X \to Y$ be continuous and $G$-equivariant,
and let $K\subset X$ be compact.  

\begin{enumerate}
\item[(i)]
The models in Eq.~\eqref{eq:neural-ansatz} can approximate $F$ uniformly on $K$
if and only if the polynomial coefficient models in Eq.~\eqref{eq:polynomial-ansatz} can do so.  These hold if and only if there exists a map $a : \mathbb{R}^R\to\mathbb{R}^M$ such that 
$$F(x)=\sum_{i=1}^M
a_i\bigl(r_1(x),\ldots,r_R(x)\bigr)m_i(x)
\qquad\text{for all }x\in K,$$
where continuity of $a$ is not required.

\item[(ii)]
If $G$ is compact and acts continuously on $X$ and $Y$,
the $r_j$ generate $\mathcal{R}(X)$ as an algebra, and the $m_i$ generate
$\mathcal{M}(X,Y)$ as an $\mathcal{R}(X)$-module, then the models in Eq.~\eqref{eq:neural-ansatz}
can approximate $F$ uniformly on $K$.
\end{enumerate}
\end{proposition}
\begin{proof}
See Appendix \ref{sec:proof-expressivity}.
\end{proof}

We now turn to the question of computing the generators
$\set{r_1,\dots,r_R}$ and $\set{m_1,\dots,m_M}$ appearing in
Eq.~\eqref{eq:neural-ansatz}. 
As an example, when $G=\Orth(d)$,
$X=n\dsR^d$, and $Y=\dsR^d$, the invariant algebra is generated by
the pairwise inner products $x_i\cdot x_j$, and the covariant module is
generated by the coordinate projections $(x_1, \ldots, x_n) \mapsto x_j$
\cite[Lemma~1 and Proposition~4]{VillarEtAl2021}. When $G=\SO(3)$,
Gordan's algorithm or more modern variants can be used to compute the generators
up to a given degree
\cite{Olive2017, KommalapatiEtAl2026}.
The e3nn
\cite{GeigerSmidt2022} and lie-nn
\cite{BatatiaEtAl2023} packages can be used to compute explicit vector space bases for the homogeneous components of the covariant module and invariant algebra for various groups. 
One can then apply the graded Nakayama lemma to extract
minimal homogeneous generators from these bases. 

One of our technical contributions is an alternative construction of vector space bases (Theorem \ref{thm:isotypic-decomposition}) that uses the Cauchy
decomposition to factor spaces tracking the action of $G$ from spaces tracking identical copies of irreducible representations in the input representation. 
This results in an algorithm (Algorithm~\ref{alg:extract-independent-generators}) to compute minimal
homogeneous generators for the invariant algebra and covariant module
up to prescribed degrees.
We also introduce data structures based on tensor networks 
to enable efficient storage and evaluation of the  invariant
and covariant features (Definition~\ref{def:isotypic-data}).
Our method achieves large speedups (Figures~\ref{fig:gordan-degree} and~\ref{fig:gordan-input}) compared to a Gordan's algorithm baseline for $\SO(3)$. %

Conceptually our main contribution is a general approach to equivariant learning that consolidates various constructions from many application areas into a common framework (Section~\ref{sec:applications}). 
We make this approach practical with a user-friendly \href{https://github.com/MichaelLiu2024/Equivariant-Polynomial-Tensor-Functions}{Python package}. 
The user declares the input and output representations, the feature degree cutoffs, the coefficient functions, and the training data; the package computes minimal homogeneous generators and fits the resulting equivariant model in a few lines of code. 
In particular, the coefficient fitting functions can be any PyTorch module or any estimator with fit and predict methods, such as those in scikit-learn. 
Our package has built-in support for $\SO(2)$, $\Orth(2)$, $\SO(3)$, $\Orth(3)$, and $\mathfrak S_n$, and it accepts other groups provided that the required representation-theoretic data are supplied.
The resulting equivariant models can be viewed as end-to-end predictors, or as layerwise components in larger architectures.

The rest of the paper is organized as follows.
Section~\ref{sec:applications} shows how existing equivariant learning models instantiate the general framework.
Section~\ref{sec:theory-and-algorithms} develops our algorithm for computing minimal homogeneous invariant and covariant generators up to prescribed degrees.
Section~\ref{sec:examples-algebra} applies the algorithm to various groups and compares it with Gordan's algorithm for $\SO(3)$.
Section~\ref{sec:nbody-results} demonstrates the full learning pipeline on the water Hamiltonian prediction problem and the EGNN benchmark.
These show the model used as an end-to-end predictor and within layers of larger architectures, respectively.
Section~\ref{sec:conclusion} summarizes our contributions and outlines future directions.
Appendix~\ref{sec:appendix-preliminaries} explains in detail how we construct the vector space bases for our algorithm.
Appendix~\ref{sec:appendix-proofs} presents proofs for our results.

\section{Examples of the model in prior works}\label{sec:applications}

\begin{example}[Equivariant multilayer perceptrons]\label{ex:application-emlp}
Finzi et al.~\cite[Section~6]{FinziWellingWilson2021} build equivariant multilayer
perceptrons (EMLPs) from affine maps, bilinear contractions, and scalar
gates. For a block between direct sums $U,V$ of tensor representations, take
\[
  G=\Orth(3),\qquad X=U,\qquad Y=V.
\]
Each gated output channel has the form
\[
  F_{\theta,a}(x)=\sigma(r_a(x))\,m_a(x),
\]
where $\sigma$ is the sigmoid, $m_a$ a polynomial covariant, and $r_a$
an invariant polynomial, both of degree at most two. Expanding $m_a$ and
$r_a$ in fixed polynomial bases expresses each block exactly in the
ansatz, with scalar coefficients given by the gates and expansion weights.
The full EMLP composes such blocks on successive hidden representations.
The EMLP constraint solver computes a basis of the equivariant linear maps, the
degree-one component of $\scM(X,Y)$, whereas
Algorithm~\ref{alg:extract-independent-generators} computes generators in all
degrees up to a prescribed bound.
\end{example}

\begin{example}[Equivariant graph neural networks]\label{ex:application-egnn}
Satorras et al.~\cite[Eqs.~(3)-(6)]{SatorrasHoogeboomWelling2021} construct equivariant graph neural network (EGNN) layers
from invariant edge messages and relative vectors. For one edge, take
\[
  G=\Orth(3),\qquad X=\dsR^k\oplus\dsR^3,
  \qquad Y=\dsR^3.
\]
Write $s_{ij}=(h_i,h_j,a_{ij})\in\dsR^k$ for scalar node and edge
features and $u_{ij}=q_i-q_j \in \mathbb{R}^3$. For $n\ge2$ nodes, the coordinate update is
\[
  q_i'=q_i+\frac1{n-1}\sum_{j: j\ne i}
    \scN_\theta(s_{ij},\lVert u_{ij}\rVert^2)\,u_{ij}.
\]
Each vector edge message is exactly the ansatz with
$r=(s_{ij},\lVert u_{ij}\rVert^2)$, covariant $u_{ij}$, and
$\scN_\theta=\phi_x\circ\phi_e$. Scalar messages use covariant $1$.
The ansatz guarantees orthogonal equivariance of each message.
Shared maps, summation, scalar node updates, and the coordinate residual
assemble the layer. 
Relabeling node and edge features gives permutation
equivariance, while relative displacements and the residual give translation
equivariance.
\end{example}

\begin{example}[Reynolds stress correction]\label{ex:application-reynolds-stress}
For turbulence modeling, Schmelzer et al.~\cite{SchmelzerDwightCinnella2020} learn a stress
anisotropy correction from normalized strain $S$ and rotation $W$. Take
\[
  G=\Orth(3),\qquad
  X=\Sym^2_0(\dsR^3)\oplus\Lambda^2(\dsR^3),
  \qquad Y=\Sym^2_0(\dsR^3).
\]
Here $S$ is symmetric and trace-free, and $W$ is skew-symmetric. For planar fluid 
flow, their model uses
\[
  \begin{gathered}
    F_\theta(S,W)=\sum_{i=1}^3\scN_{\theta,i}(r)\,m_i,
    \qquad r=(\tr(S^2),\tr(W^2)),\\[3pt]
    m_1=S,\qquad m_2=SW-WS,\qquad
    m_3=S^2-\tfrac13\tr(S^2)I_3.
  \end{gathered}
\]
The $m_i$ are polynomial covariants. Sparse regression selects each
coefficient from the span of 16 prescribed monomials in $r$.
\end{example}

\begin{example}[Sparse vector recovery]\label{ex:application-sparse-recovery}
For data science applications, Gregory et al.~\cite{GregoryEtAl2026} recover a sparse vector in a
subspace with orthonormal basis $B\in\dsR^{n\times d}$. Take
\[
  G=\Orth(d),\qquad X=n\dsR^d,\qquad
  Y=\Sym^2(\dsR^d).
\]
Basis changes act by $B\mapsto BQ^{\top}$ and by conjugation on $Y$.
Their predictor has the equivalent matrix form
\[
  F_\theta(B)=\scN_{\theta,0}(r)I_d+B^{\top}\scN_{\theta,1}(r)B,
  \qquad r=BB^{\top},
\]
where $\scN_{\theta,0}$ is scalar-valued and $\scN_{\theta,1}$ returns a symmetric $n\times n$
coefficient matrix. The covariants are the identity and symmetrized row
outer products. 
A unit leading eigenvector $u$ gives the estimate $Bu$.
\end{example}

\begin{example}[Gradient-domain kernel potentials]\label{ex:application-sgdml}
In molecular dynamics, the sGDML model~\cite{ChmielaEtAl2018} learns conservative molecular
forces from force observations. For its energy on collision-free
configurations $q=(q_i)$, take
\[
  G=\Orth(3),\qquad X=n\dsR^3,\qquad Y=\dsR.
\]
Squared distances give polynomial invariants:
\[
  r(q)=\bigl(\lVert q_i-q_j\rVert^2\bigr)_{i<j},\qquad
  E_\theta(q)=\scN_\theta(r(q)),\qquad f_\theta(q)=-\nabla_qE_\theta(q).
\]
The coefficient function is a permutation-symmetrized kernel expansion
in inverse distances $r_{ij}^{-1/2}$, including derivatives with respect
to fixed training configurations. Thus the energy uses covariant $1$.
The gradient readout gives conservative, translation-invariant,
orthogonally equivariant forces.
\end{example}

\begin{example}[Atomic cluster expansion]\label{ex:application-ace}
In computational physics, the atomic cluster expansion (ACE)~\cite{KovacsEtAl2021} predicts site
energies from pooled radial and spherical harmonic channels. Take
\[
  G=\Orth(3),\qquad X=\bigoplus_{b,a,\ell}H_\ell,
  \qquad Y=\dsR,
\]
where $H_\ell$ is the degree-$\ell$ harmonic representation. For nonzero
neighbor displacements $u_j$, the channels and energy are
\[
  \begin{gathered}
    A_{ba\ell}=\sum_{j\in b}R_{a\ell}(\lVert u_j\rVert)
      Y_\ell(\widehat u_j),
    \qquad \widehat u_j=\frac{u_j}{\lVert u_j\rVert},\\[3pt]
    E_\theta(A)=\sum_\alpha c_{\theta,\alpha} B_\alpha(A).
  \end{gathered}
\]
Here $b,a$ label species and radial channels, and $Y_\ell$ is a $(2 \ell + 1)$-dimensional
spherical harmonic vector. The invariant polynomial contractions $B_\alpha$ give
$r$, with covariant $1$ and linear coefficient function. Pooling enforces
permutation invariance within species. %
\end{example}

\section{Minimal homogeneous generators for the invariant algebra and covariant 
module}\label{sec:theory-and-algorithms}

In this section, we present our algorithm to compute minimal homogeneous generators for the invariant algebra and covariant module up to given degrees.
Let $G$ be an arbitrary group, let
$k\in\{\dsR,\dsC\}$, and let $X,Y$ be finite-dimensional semisimple $G$-representations over $k$.
We first recall basic facts.

\begin{lemma}\label{lem:semisimple-constructions}
The collection of finite-dimensional semisimple representations is closed under tensor products, duals, subrepresentations, quotients, and Schur functors.\footnote{Schur functors generalize symmetric and exterior powers.  See Appendix~\ref{sec:appendix-preliminaries} for more.}
\end{lemma}
\begin{proof}
See Chevalley's theorem \cite[Theorem~4.12.1]{EGNO2015}.
\end{proof}

\begin{lemma}\label{lem:invariant-finite-generation}
The invariant algebra $\scR(X)$ is finitely generated over $k$.
\end{lemma}
\begin{proof}
See Hilbert's finiteness theorem \cite[\S1.6, Theorem~1]{KraftProcesi1996}.
\end{proof}

\begin{lemma}\label{lem:covariant-finite-generation}
The covariant module $\scM(X,Y)$ is finitely generated over $\scR(X)$.
\end{lemma}
\begin{proof}
Note $\scM(X,Y)$ identifies with the subset of $\mathcal{R}(X \oplus Y^*)$ consisting of polynomials that are homogeneous of degree $1$ in the $Y^*$ variables, while $\mathcal{R}(X)$ consists of elements of $\mathcal{R}(X \oplus Y^*)$ that are degree $0$ with respect to $Y^*$.
By Lemma~\ref{lem:invariant-finite-generation}, $\mathcal{R}(X \oplus Y^*)$ has a finite generating set homogeneous in the \(Y^*\) variables. 
The degree one generators generate the degree one part over the degree zero part.
\end{proof}

For simplicity, we develop our algorithm over $k=\dsC$. The  reason is Schur's lemma, which says that $\End_G(H)=\dsC$ for every irreducible representation $H$. When $k = \dsR$, the essential change is that multiplicity spaces become modules over the division algebras \(\operatorname{End}_G(H) \cong \dsR, \dsC, \dsH\), and compositions through irreducibles use tensor products over these algebras. 
One can obtain variants of Theorem \ref{thm:isotypic-decomposition} and Algorithm \ref{alg:extract-independent-generators} below over \(k=\mathbb R\). However instead of doing this, we cite a standard result in Proposition \ref{prop:complexification-descent} to recover the real case from the complex case.

Splitting the covariant module and invariant algebra into homogeneous pieces by total polynomial degree, we have
\[
  \scM(X,Y)=\bigoplus_{D\ge0}\scM^D(X,Y),\qquad
  \scM^D(X,Y)\coloneq\HomG{\Sym^D X}{Y},
\]
and
\[
  \scR(X)=\bigoplus_{D\ge0}\scR^D(X),\qquad
  \scR^D(X)\coloneq\HomG{\Sym^D X}{\mathbf1}.
\]
Here $\operatorname{Sym}^D(\cdot)$ denotes the $D$-th symmetric power, and $\mathbf1 \cong k$ is the trivial representation.  Write $\operatorname{Irr}(G)$ for the set of isomorphism classes of finite-dimensional complex irreducible representations of $G$, and for each $\lambda\in\operatorname{Irr}(G)$, fix a representative $H_\lambda$ of the class $\lambda$. Let
\[
  X\cong\bigoplus_\lambda H_\lambda\otimes\HomG{H_\lambda}{X},
  \qquad
  Y\cong\bigoplus_\nu H_\nu\otimes\HomG{H_\nu}{Y}
\]
be the isotypic decompositions \cite[Proposition~4.1.15]{GW2009} of $X$ and $Y$.

It suffices to consider the case where the output representation $Y = H_\nu$ is irreducible.
Indeed, applying $\HomG{\Sym^D X}{-}$ to the isotypic decomposition of $Y$ gives an isomorphism
\begin{equation}\label{eq:output-reduction}
  \bigoplus_{\nu}
  \HomG{\Sym^D X}{H_\nu}\otimes\HomG{H_\nu}{Y}
  \xrightarrow{\ \sim\ }
  \HomG{\Sym^D X}{Y}, %
\end{equation}
which allows us to pass vector space bases for $\HomG{\Sym^D X}{H_\nu}$ to those for $\HomG{\Sym^D X}{Y}$. Thus, we begin the algorithm development by computing a vector space basis for $\HomG{\Sym^D X}{H_\nu}$.

\begin{table}[htbp]
\centering
\small
\setlength{\tabcolsep}{5pt}
\begin{tabular}{>{\raggedright\arraybackslash}p{0.20\textwidth}
>{\raggedright\arraybackslash}p{0.30\textwidth}
>{\raggedright\arraybackslash}p{0.30\textwidth}}
\toprule
  & $Y$ irreducible & $Y$ arbitrary \\
\midrule
Vector space basis
&
Algorithm~\ref{alg:stream-vector-space-basis}
&
Proposition~\ref{prop:arbitrary-output}
  \\[0.5em]
Minimal homogeneous generators
&
Algorithm~\ref{alg:extract-independent-generators}
&
Proposition~\ref{prop:arbitrary-output}
  \\
\bottomrule
\end{tabular}
\caption{Roadmap for Section \ref{sec:theory-and-algorithms}.
We start in the top left corner.
We use the graded Nakayama lemma to move down the table.
We use Eq.~\eqref{eq:output-reduction} to move right in the table.}
\label{tab:main-results-guide}
\end{table}

\subsection{Vector space bases}\label{sec:vector-space-bases}

As vector spaces, a homogeneous piece of the invariant algebra is just $\HomG{\Sym^D X}{H_\nu}$, where $H_\nu = \mathbf 1$ is the trivial representation. Thus, we consider only covariants below.  In the next theorem, we use the Cauchy decomposition to separate repeated copies of each irreducible input type from equivariant maps coupling the irreducible representations.

\begin{theorem}[Isotypic decomposition of $\Sym^D X$]\label{thm:isotypic-decomposition}
Let $\nu\in\operatorname{Irr}(G)$ and $D\in\dsN_0$. Let $\lambda$ index the irreducible labels occurring in $X$.
Let
\begin{itemize}[leftmargin=2.5em]
  \item $(d_\lambda)_\lambda$ range over tuples of nonnegative integers with $\sum_\lambda d_\lambda=D$
  \item $(\pi_\lambda)_\lambda$ range over tuples of partitions $\pi_\lambda\vdash d_\lambda$ with at most
  $\min\{\dim H_\lambda,\dim\HomG{X}{H_\lambda}\}$ parts
  \item $(\mu_\lambda)_\lambda$ range over tuples of irreducible labels with $\mu_\lambda$ occurring in
  $\mathbf S_{\pi_\lambda}(H_\lambda)$ for every $\lambda$ and
  $\nu$ occurring in $\bigotimes_\lambda H_{\mu_\lambda}$.
\end{itemize}
Let
\begin{itemize}[leftmargin=2.5em]
  \item $C$ range over a basis for $\HomG{\bigotimes_\lambda H_{\mu_\lambda}}{H_\nu}$
  \item $L_\lambda$ range over a basis for $\HomG{\mathbf S_{\pi_\lambda}(H_\lambda)}{H_{\mu_\lambda}}$ for each $\lambda$
  \item $T_\lambda$ range over a basis for $\mathbf S_{\pi_\lambda}\!\left(\HomG{X}{H_\lambda}\right)$ for each $\lambda$.
\end{itemize}
Then the maps $Q$ defined by
\begin{equation}\label{eq:isotypic-evaluation}
  Q(x^{\otimes D})
  \coloneq C\!\left(\bigotimes_\lambda L_\lambda(T_\lambda(x))\right)
\end{equation}
form a basis for $\HomG{\Sym^D X}{H_\nu}$, where $T_\lambda(x)$ is defined in
Eq.~\eqref{eq:appendix-tableau-input}.
\end{theorem}
\begin{proof}[Sketch]
The claim follows by transporting product bases through the isomorphism
\begin{multline}\label{eq:isotypic-decomposition}
\bigoplus_{(d_\lambda)_\lambda}\bigoplus_{(\pi_\lambda)_\lambda}\bigoplus_{(\mu_\lambda)_\lambda}\HomG{\bigotimes_\lambda H_{\mu_\lambda}}{H_\nu}\otimes\bigotimes_\lambda\HomG{\mathbf S_{\pi_\lambda}(H_\lambda)}{H_{\mu_\lambda}}\otimes\bigotimes_\lambda \mathbf S_{\pi_\lambda}\!\left(\HomG{X}{H_\lambda}\right)
\\
\xrightarrow{\ \sim\ }\HomG{\Sym^D X}{H_\nu},
\end{multline}
which sends $C\otimes\bigotimes_\lambda L_\lambda\otimes\bigotimes_\lambda T_\lambda$
to the map $Q$ in Eq.~\eqref{eq:isotypic-evaluation}.
See Appendix~\ref{sec:proof-main-decomposition} for the full proof.
\end{proof}

\begin{definition}\label{def:isotypic-data}
We represent a covariant $Q$ as in Eq.~\eqref{eq:isotypic-evaluation} via its \emph{isotypic data}:
\begin{itemize}[leftmargin=2.5em]
  \item a tensor train representing $C$ (see Appendix~\ref{sec:prelim-tensor-trains})
  \item for each $\lambda$, a tensor tree representing $L_\lambda$ (see Appendix~\ref{sec:appendix-leaf-bases})
  \item for each $\lambda$, a semistandard Young tableau representing $T_\lambda$ (see Appendix~\ref{sec:appendix-schur-evaluation}).
\end{itemize}
\end{definition}

\begin{figure}[H]
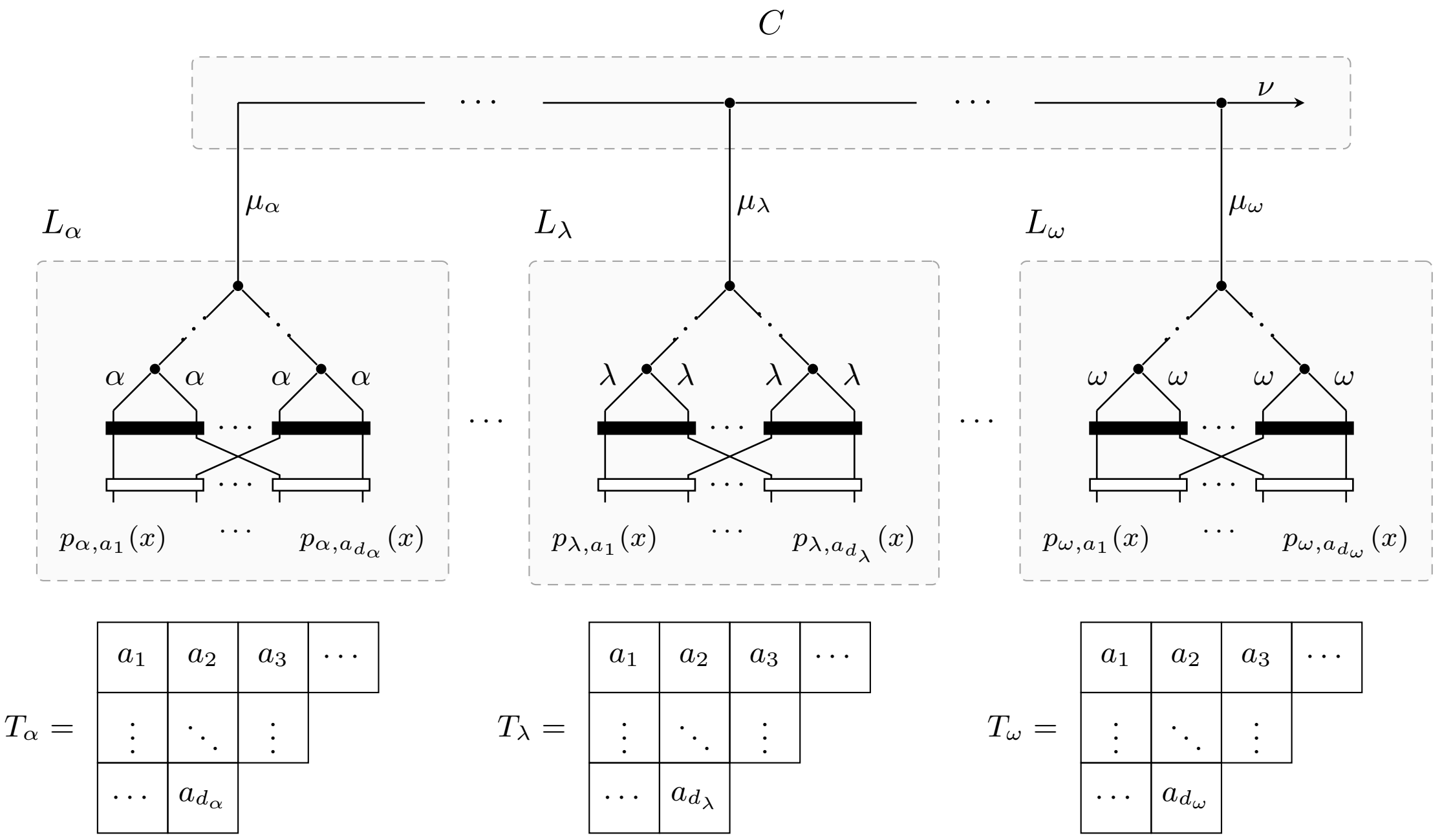

\centering
\IsoAbstractDiagram
\caption{Isotypic data for an arbitrary covariant $Q$ as in Definition \ref{def:isotypic-data}. Each $\set{p_{\lambda,a}}_a$ is a basis for the multiplicity space $\HomG{X}{H_\lambda}$ in the isotypic decomposition of $X$, and the $\alpha,\ldots,\lambda,\ldots,\omega$ label the irreducible components of $X$.
Empty and filled bars denote row symmetrization and column
antisymmetrization respectively \cite[Sections~6.1-6.2 and~9.4]{Cvitanovic2008}.}
\label{fig:abstract-isotypic-data}
\end{figure}

\begin{figure}[H]
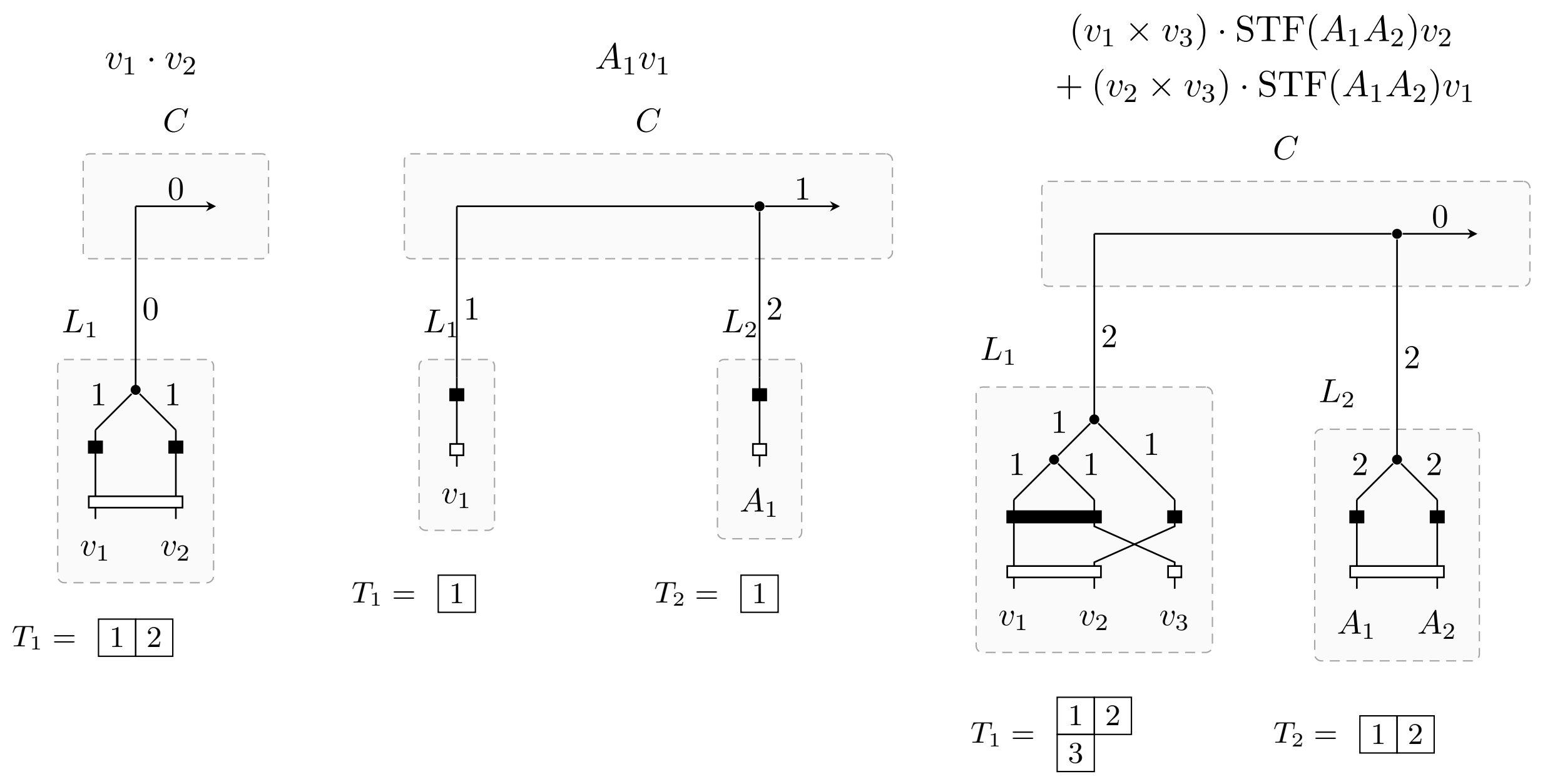

\centering
\begin{minipage}[t]{.22\linewidth}\centering
  \strut$v_1\cdot v_2$\par\medskip
  \begin{isotypicdiagram}[baseline=(iso-core-frame.north),iso tableau rows=1]
    \IsoLeaf[copies={v_1,v_2}]{v}{0}{1}{{1,2}}{
      \IsoFuse{r}{s1}{s2}{0}
    }
    \IsoCore{v}{0}
  \end{isotypicdiagram}
\end{minipage}\hfill
\begin{minipage}[t]{.36\linewidth}\centering
  \strut$A_1v_1$\par\medskip
  \begin{isotypicdiagram}[baseline=(iso-core-frame.north),iso tableau rows=1]
    \IsoLeaf[copies={v_1}]{v}{0}{1}{{1}}{}
    \IsoLeaf[copies={A_1}]{a}{3.2}{2}{{1}}{}
    \IsoCore{v,a}{1}
  \end{isotypicdiagram}
\end{minipage}\hfill
\begin{minipage}[t]{.40\linewidth}\centering
  \strut$\begin{gathered}
    (v_1\times v_3)\cdot\operatorname{STF}(A_1A_2)v_2\\
    {}+(v_2\times v_3)\cdot\operatorname{STF}(A_1A_2)v_1
  \end{gathered}$\par\medskip
  \begin{isotypicdiagram}[baseline=(iso-core-frame.north),iso tableau rows=2]
    \IsoLeaf[copies={v_1,v_2,v_3}]{v}{0}{1}{{1,2},{3}}{
      \IsoFuse{column}{s1}{s3}{1}
      \IsoFuse{r}{column}{s2}{2}
    }
    \IsoLeaf[copies={A_1,A_2}]{a}{3.2}{2}{{1,2}}{
      \IsoFuse{r}{s1}{s2}{2}
    }
    \IsoCore{v,a}{0}
  \end{isotypicdiagram}
\end{minipage}
\caption{Isotypic data for the dot product, the matrix-vector product, and a fifth degree covariant.}
\label{fig:so3-dot-matrix-vector}
\label{fig:so3-cross-contraction}
\end{figure}

\begin{example}\label{ex:isotypic-data-so3}
Let $G=\SO(3)$, $H_0=\dsC$, $H_1=\dsC^3$, and $H_2 = \Sym_0^2(\dsC^3)$.
Write $\operatorname{STF}$ for symmetrization and trace removal, i.e., $\operatorname{STF}(A) = \tfrac{1}{2}(A + A^{\top}) - \tfrac{1}{3}\operatorname{tr}(A) I_3$ for $A \in \mathbb{C}^{3 \times 3}$.
The dot product, the matrix-vector product, and a fifth degree covariant are
represented up to normalization by the following \textit{tensor networks} (see Appendix \ref{sec:prelim-tensor-trains}), where edges carry irreducible representation labels and nodes carry Clebsch-Gordan tensors:
\begin{center}
\begin{tensordiagram}[baseline=(current bounding box.north)]
  \TensorInput{v1}{0}{v_1}{1}
  \TensorInput{v2}{1}{v_2}{1}
  \IsoFuse{root}{v1}{v2}{0}
  \TensorOutput{v_1\cdot v_2}
\end{tensordiagram}
\hspace{1.7cm}
\begin{tensordiagram}[baseline=(current bounding box.north)]
  \TensorInput{v1}{0}{v_1}{1}
  \TensorInput{a1}{1}{A_1}{2}
  \IsoFuse{root}{v1}{a1}{1}
  \TensorOutput{A_1v_1}
\end{tensordiagram}
\hspace{1.7cm}
\begin{tensordiagram}[baseline=(current bounding box.north)]
  \TensorInput{v1}{0}{v_1}{1}
  \TensorInput{v2}{1}{v_2}{1}
  \TensorInput{v3}{2}{v_3}{1}
  \TensorInput{a1}{3}{A_1}{2}
  \TensorInput{a2}{4}{A_2}{2}
  \IsoFuse{cross}{v1}{v2}{1}
  \IsoFuse{vectors}{cross}{v3}{2}
  \IsoFuse{matrices}{a1}{a2}{2}
  \IsoFuse{root}{vectors}{matrices}{0}
  \TensorOutput{(v_1\times v_2)\cdot\operatorname{STF}(A_1A_2)v_3}
\end{tensordiagram}
\end{center}

Figure~\ref{fig:so3-dot-matrix-vector} displays the isotypic data for the first two maps and a related fifth-degree covariant.

\end{example}

The next result gives a very rough bound on the complexity for storage and evaluation of a covariant using the data structure of its isotypic data.  The point is that for fixed $n$ the costs with respect to the degree $D$ are polynomial, and not exponential.  More informative bounds are given in Appendix~\ref{sec:proof-isotypic-complexity}.

\begin{proposition}
\label{prop:isotypic-complexity}
Let $Q$ be a covariant of degree $D \geq 1$ represented by isotypic data
as in Definition~\ref{def:isotypic-data}, and let $n=\dim X\ge1$.
Assume that all required Clebsch-Gordan tensors are precomputed and included
in storage. Then the isotypic data occupy $O((D+2)^{2n})$ numerical entries in storage,
and the procedure in Appendix~\ref{sec:appendix-evaluation} evaluates $Q(x^{\otimes D})$
using $O((D+2)^{2n})$ arithmetic operations. Implied constants are absolute.
\end{proposition}
\begin{proof}
See Appendix~\ref{sec:proof-isotypic-complexity}.
\end{proof}

To implement concrete algorithms, we henceforth assume access to the information in Assumption~\ref{ass:computational} below. The Macaulay2 packages InvariantRing \cite{FerraroEtAl2024} and SchurRings \cite{SchurRings2026} support invariant ring and character computations; explicit multiplicity space bases are additional inputs to our algorithms.

\begin{assumption}\label{ass:computational}
Our algorithms have access to
\begin{itemize}[leftmargin=2em]
  \item explicit bases for the multiplicity spaces $\HomG{X}{H_\lambda}$ and $\HomG{H_\nu}{Y}$ appearing in the isotypic decompositions of $X$ and $Y$
  \item explicit bases for the multiplicity spaces $\HomG{H_\alpha\otimes H_\beta}{H_\gamma}$ appearing in the Clebsch-Gordan decompositions 
  \item the irreducible constituents of the Schur powers
  $\mathbf S_\pi(H_\lambda)$ along with their (positive integer) multiplicities.
\end{itemize}
\end{assumption}

Knowledge of the Schur multiplicities enables early
termination in Step \ref{alg:stream-vector-space-basis:step-schur-tensor-tree-basis} of Algorithm \ref{alg:stream-vector-space-basis} below. 
While the Schur multiplicities can also be obtained directly by computing the rank of the spanning tensor trees in
Appendix~\ref{sec:appendix-leaf-bases}, this can substantially
increase the computational cost.

Constructing the full basis in Theorem \ref{thm:isotypic-decomposition} may be impractical when $D$ is large.
Importantly, for applications such as the computation of minimal homogeneous generators, it suffices to process a random stream from this basis.

\begin{algorithm}[\texttt{stream\_vector\_space\_basis}]\label{alg:stream-vector-space-basis}

\phantom{}

\begin{enumerate}[label=\textbf{Step \arabic*.},ref=\arabic*,leftmargin=2.8cm]

\item[\textbf{Input:}] bases for the multiplicity spaces $\HomG{X}{H_\lambda}$ appearing in the isotypic decomposition of $X$, an output irreducible representation label $\nu\in\operatorname{Irr}(G)$, and a degree $D \in \dsN_0$.

\item[\textbf{Output:}] bases for the tensor factors of a randomly chosen direct summand in Eq.~\eqref{eq:isotypic-decomposition}.

\item \label{alg:stream-vector-space-basis:step-summand} Randomly choose a direct summand in Eq.~\eqref{eq:isotypic-decomposition} (without replacement).

\item \label{alg:stream-vector-space-basis:step-bases} Return
\begin{enumerate}[label=\textup{(\alph*)},ref=\theenumi(\alph*),leftmargin=2.2em]
  \item \label{alg:stream-vector-space-basis:step-tensor-train-basis} a tensor train basis for $\HomG{\bigotimes_{\lambda} H_{\mu_\lambda}}{H_\nu}$ (see Appendix~\ref{sec:prelim-tensor-trains}) 
  \item \label{alg:stream-vector-space-basis:step-schur-tensor-tree-basis} a tensor tree basis for each $\HomG{\mathbf S_{\pi_\lambda}(H_\lambda)}{H_{\mu_\lambda}}$ (see Appendix~\ref{sec:appendix-leaf-bases}) 
  \item \label{alg:stream-vector-space-basis:step-ssyt-basis} a semistandard Young tableaux basis for each $\mathbf S_{\pi_\lambda}\!\left(\HomG{X}{H_\lambda}\right)$ (see Appendix~\ref{sec:appendix-schur-evaluation}).
\end{enumerate}
\end{enumerate}
\end{algorithm}

\subsection{Minimal homogeneous generators}\label{sec:minimal-homogeneous-generators}

Here, we use the graded Nakayama lemma to extract minimal homogeneous generators from the random stream obtained above. Write $\scR_+(X)\coloneq\bigoplus_{D>0}\scR^D(X)$. Since $\scR^0(X)=\dsC$, the graded Nakayama lemma \cite[Lemma~3.5.1]{DerksenKemper2002} \cite[Section~2]{Domokos2020} implies the following criteria for homogeneous $L\subseteq\scM(X,Y)$ in the module case and $L\subseteq\scR_+(X)$ in the algebra case:
\begin{equation}\label{eq:graded-nakayama-criteria}
  \begin{aligned}
    L\text{ minimally generates }\scM(X,Y)\text{ as an }\scR(X)\text{-module}
    &\iff
    \overline L\text{ is a basis for }
    \scM(X,Y)\big/\scR_+(X)\scM(X,Y),\\
    L\text{ minimally generates }\scR(X)\text{ as a }\dsC\text{-algebra}
    &\iff
    \overline L\text{ is a basis for }
    \scR_+(X)\big/\scR_+(X)^2.
  \end{aligned}
\end{equation}
The resulting method to extract minimal homogeneous generators is standard \cite[Theorem~6.1 and Algorithm~6.4]{HubertRodriguezBazan2022} \cite[Section~3.3 and Appendix~F]{KommalapatiEtAl2026}, and we record a simplified version of our variant.

\begin{algorithm}[\texttt{extract\_independent\_generators}]\label{alg:extract-independent-generators}
\phantom{}
\begin{enumerate}[label=\textbf{Step \arabic*.},leftmargin=2.8cm]

\item[\textbf{Input:}] bases for the multiplicity spaces
$\HomG{X}{H_\lambda}$ appearing in the isotypic decomposition of $X$,
an output irreducible representation label $\nu\in\operatorname{Irr}(G)$, and a degree cutoff
$D_{\max}\in\dsN_0$.

\item[\textbf{Output:}] minimal homogeneous generators through degree
$D_{\max}$ for
\[
  \begin{cases}
    \scM(X,H_\nu)\text{ as an }\scR(X)\text{-module},
      & H_\nu\neq\mathbf 1,\\
    \scR(X)\text{ as a }\dsC\text{-algebra},
      & H_\nu=\mathbf 1.
  \end{cases}
\]

\item Initialize $L_0=\emptyset$.

\item For each degree $D=1,\ldots,D_{\max}$ in increasing order:
\begin{enumerate}[label=\textup{(\alph*)},leftmargin=2.2em]
\item Set
\[
  I_D\coloneq
  \begin{cases}
    \{0,\ldots,D-1\},
    & H_\nu\neq\mathbf 1,\\
    \{1,\ldots,\lfloor D/2\rfloor\},
    & H_\nu=\mathbf 1.
  \end{cases}
\]

\item Run
Algorithm~\ref{alg:stream-vector-space-basis} on $H_\nu$
at degree $D$ to initialize a random stream from
$\scM^D(X,H_\nu)$. For each $d\in I_D$, repeatedly run Algorithm~\ref{alg:stream-vector-space-basis} on $\mathbf 1$ at degree $D-d$
to obtain a basis $R_{D-d}$ for
$\scR^{D-d}(X)$. Compute
$N_D\coloneq\dim\scM^D(X,H_\nu)$ using
Theorem~\ref{thm:isotypic-decomposition}.%

\item Set
\[
  P_D\coloneq\{rg\mid d\in I_D,\ r\in R_{D-d},\ g\in L_d\}.
\]

\item Set
\[
  m\coloneq\left\lceil\frac{N_D}{\dim H_\nu}\right\rceil+2.
\]
Upon retry, replace $m$ by $m+1$.

\item Choose random probes $x_1,\ldots,x_m\in X$. For any $f\in\scM^D(X,H_\nu)$, define the evaluation column
\[
  E(f)\coloneq
  \begin{bmatrix}f(x_1)\\ \vdots\\ f(x_m)\end{bmatrix}
  \in\dsC^{m\dim H_\nu}.
\]

\item Set $L_D=\emptyset$. Initialize $E$ as a maximal linearly
independent subset of $\{E(p):p\in P_D\}$.

\item For each candidate $f$ in the random stream, if $E(f)$ increases the rank of $E$, append $E(f)$ to $E$
and $f$ to $L_D$; otherwise, discard $f$. Continue until $\operatorname{rank}E=N_D$ or the stream is exhausted.

\item If $\operatorname{rank}E<N_D$ and the stream is exhausted, retry steps (b)-(g). Record $L_D$ once $\operatorname{rank}E=N_D$, or report failure after too many retries.
\end{enumerate}

\item Return $L_0, L_1, \ldots, L_{D_{\max}}$.
\end{enumerate}
\end{algorithm}

The core idea behind Algorithm \ref{alg:extract-independent-generators} is as follows. Since
\[
  \operatorname{span}P_D
  =\sum_{d\in I_D}\scR^{D-d}(X)\operatorname{span}L_d
  =\begin{cases}
    \bigl(\scR_+(X)\scM(X,H_\nu)\bigr)^D,
      & H_\nu\neq\mathbf 1\\
    \bigl(\scR_+(X)^2\bigr)^D,
      & H_\nu=\mathbf 1,
  \end{cases}
\]
it follows from Eq.~\eqref{eq:graded-nakayama-criteria} that extending a basis for $\operatorname{span} P_D$ via $L_D$ to a basis for $\scM^D(X,H_\nu)$ indeed gives minimal homogeneous generators in degree $D$.

Our implementation of Algorithm \ref{alg:extract-independent-generators} processes candidate generators by multidegree $(d_\lambda)_\lambda$ instead of total degree $D$, which reduces the size of the evaluation matrix $E$.
Moreover, we evaluate candidates in the batches returned by Algorithm \ref{alg:stream-vector-space-basis}, which enables more efficient tensor contractions.
Finally, when all Clebsch-Gordan tensors involved have integer entries, we use finite field evaluation and rank selection to avoid numerical instability at high degrees.
With a sufficiently large prime $p$ and sufficiently many independent uniform probes, rank selection is correct with high probability~\cite[Lemmas~1-2]{Schwartz1980}. We use $p=65521$ as a heuristic choice. %

\subsection{Extensions of the basic construction}
\label{sec:arbitrary-outputs}
\label{sec:index-two-extensions}
\label{sec:complexification-real-descent}
\label{sec:copy-permutation-invariance}

The following results extend the above constructions to arbitrary output representations,
group extensions of index two, permutation invariance, and real representations.

\begin{proposition}[Bases and generators for arbitrary output representations]\label{prop:arbitrary-output}
Let $\nu$ index the irreducible components of $Y$. A basis for $\scM^D(X,Y)$ can be obtained by composing bases for
$\scM^D(X,H_\nu)$ with bases for $\HomG{H_\nu}{Y}$. Moreover, minimal module generators for $\scM(X,Y)$ in degree $D$
can be obtained by composing minimal module generators
for $\scM(X,H_\nu)$ in degree $D$ with bases for $\HomG{H_\nu}{Y}$.
\end{proposition}
\begin{proof}
Eq.~\eqref{eq:output-reduction} proves the basis statement.
For the generator statement, summing Eq.~\eqref{eq:output-reduction} over all degrees gives an
$\scR(X)$-linear graded isomorphism
\[
  \bigoplus_\nu\scM(X,H_\nu)\otimes\HomG{H_\nu}{Y}
  \xrightarrow{\ \sim\ }\scM(X,Y),
  \qquad f\otimes\iota\longmapsto\iota\circ f,
\]
where the multiplicity spaces $\HomG{H_\nu}{Y}$ are in degree zero. Passing to quotients gives
\[
  \bigoplus_\nu
  \frac{\scM(X,H_\nu)}{\scR_+(X)\scM(X,H_\nu)}
       \otimes\HomG{H_\nu}{Y}
  \xrightarrow{\ \sim\ }
  \frac{\scM(X,Y)}{\scR_+(X)\scM(X,Y)},
  \qquad [f]\otimes\iota\longmapsto[\iota\circ f],
\]
and the generator statement follows from Eq.~\eqref{eq:graded-nakayama-criteria}.
\end{proof}
For $Y = \mathbf 1$, the covariant module is
$\mathcal R(X)$ itself, with the single degree zero generator $1$.

\begin{proposition}[Generators for group extensions of index two]\label{prop:index-two-generators}
Let $\widetilde G\supset G$ be an extension of index two to which the
actions on $X$ and $Y$ extend.
Suppose that $\{r_i\}$ generates $\scR_G(X)$ as a $\dsC$-algebra, and $\{m_k\}$ generates $\scM_G(X,Y)$ as a $\scR_G(X)$-module.
For any $s\in\widetilde G\setminus G$, define
\[
  r_i^\pm=\frac{r_i\pm sr_i}{2},
  \qquad
  m_k^\pm=\frac{m_k\pm sm_k}{2}.
\]
Then $\{r_i^+\}\cup\{r_i^-r_j^-:i\le j\}$ generates
$\scR_{\widetilde G}(X)$ as a $\dsC$-algebra, and
$\{m_k^+\}\cup\{r_i^-m_k^-\}$ generates
$\scM_{\widetilde G}(X,Y)$ as a $\scR_{\widetilde G}(X)$-module.
\end{proposition}
\begin{proof}
Since $s^2\in G$, the projections $(1\pm s)/2$ are available without assuming compactness of $G$.
Thus, the proofs of \cite[Theorems~3.2 and~3.5]{BaptistelliManoel2013} for the algebra
statement and \cite[Theorem~4.5]{BaptistelliManoel2018} for the module
statement apply.
\end{proof}

\begin{proposition}[Bases for invariance under permutations of irreducible representations]\label{prop:copy-invariant-bases}
Let $m_\lambda=\dim\HomG{X}{H_\lambda}$, and let
$\Pi=\prod_\lambda\mathfrak S_{m_\lambda}$ independently permute the copies
of each $H_\lambda$ in $X$ and act trivially on $H_\nu$.
A basis for $\scM^D_{G\times\Pi}(X,H_\nu)$ is obtained from
Algorithm~\ref{alg:stream-vector-space-basis} by replacing each tableau
basis in Step~\ref{alg:stream-vector-space-basis:step-ssyt-basis} with a basis for
\[
  \mathbf S_{\pi_\lambda}\!\left(\HomG{X}{H_\lambda}\right)
    ^{\mathfrak S_{m_\lambda}}.
\]
In particular, the tensor train and tensor tree bases in Steps~\ref{alg:stream-vector-space-basis:step-tensor-train-basis} and~\ref{alg:stream-vector-space-basis:step-schur-tensor-tree-basis} are unchanged.
\end{proposition}
\begin{proof}
The $G$- and $\Pi$-actions commute, so
$\scM^D_{G\times\Pi}(X,H_\nu)=\scM^D_G(X,H_\nu)^\Pi$.
In Eq.~\eqref{eq:isotypic-decomposition}, $\Pi$ acts independently on
the tableau factors and trivially on the remaining factors.
\end{proof}

\begin{proposition}[Bases for invariance under permutations of identical inputs]\label{prop:diagonal-copy-invariant-bases}
Let $\mathfrak S_m$ permute the copies of $X$ in $X\otimes\dsC^m$
and act trivially on $H_\nu$.
A basis for $\scM^D_{G\times\mathfrak S_m}(X\otimes\dsC^m,H_\nu)$ is obtained from
Algorithm~\ref{alg:stream-vector-space-basis} by replacing the tensor product
of tableau bases in Step~\ref{alg:stream-vector-space-basis:step-ssyt-basis} with a basis for
\[
  \left[
    \bigotimes_\lambda
      \mathbf S_{\pi_\lambda}\!\left(\HomG{X\otimes\dsC^m}{H_\lambda}\right)
  \right]^{\mathfrak S_m}.
\]
In particular, the bases in Steps~\ref{alg:stream-vector-space-basis:step-tensor-train-basis} and~\ref{alg:stream-vector-space-basis:step-schur-tensor-tree-basis} are unchanged.
\end{proposition}
\begin{proof}
In Eq.~\eqref{eq:isotypic-decomposition} for $X\otimes\dsC^m$,
$\mathfrak S_m$ acts diagonally on the tableau factors and trivially
on the remaining factors.
\end{proof}

Our implementation computes the fixed spaces in Propositions \ref{prop:copy-invariant-bases} and \ref{prop:diagonal-copy-invariant-bases} using Reynolds averaging in orbit sum coordinates \cite[Section~3]{BergeronEtAl2008}.

\begin{proposition}[Generators for real representations]\label{prop:complexification-descent}
Let $X$ and $Y$ be real $G$-representations, and let $X_\dsC$ and $Y_\dsC$ be their complexifications. If $\{r_i\}$ generates
$\scR(X_{\dsC})$ as a $\dsC$-algebra, and $\{m_j\}$ generates
$\scM(X_{\dsC},Y_{\dsC})$ as an $\scR(X_{\dsC})$-module,
then $\{\operatorname{Re}(r_i|_X),\operatorname{Im}(r_i|_X)\}_i$ generates
$\scR(X)$ as an $\dsR$-algebra, and
$\{\operatorname{Re}(m_j|_X),\operatorname{Im}(m_j|_X)\}_j$ generates
$\scM(X,Y)$ as an $\scR(X)$-module.  These generators need not be minimal, but if  $r_i$ and $m_j$
are homogeneous then minimal real generating subsets can be
obtained using Eq.~\eqref{eq:graded-nakayama-criteria} over $\mathbb R$.
\end{proposition}
\begin{proof}
See \cite[\S1.4 and \S1.5, Exercise~29(a-b)]{KraftProcesi1996}.
\end{proof}

\section{Invariant and covariant generators for specific groups}\label{sec:examples-algebra}

We illustrate the theory and algorithms in Section \ref{sec:theory-and-algorithms} for some of the groups that arise in applications,
and include a comparison with Gordan's algorithm for $\SO(3)$.
Code to reproduce all experiments is provided in \texttt{docs/examples.ipynb}
in our \href{https://github.com/MichaelLiu2024/Equivariant-Polynomial-Tensor-Functions}{Github repository}.

\subsection{Finite abelian groups and \texorpdfstring{$\SO(2)$}{SO(2)}}\label{sec:examples-finite-abelian}\label{sec:examples-so2}

Let $G$ be a finite abelian group or $\SO(2)$, and write its character
group $\widehat G$ additively. The complex irreducible representations of
$G$ are $H_\lambda=\dsC$, $\lambda\in\widehat G$, with action
$g\cdot z=\lambda(g)z$.

Theorem~\ref{thm:isotypic-decomposition} analytically recovers the classical description of invariants and covariants as monomials of fixed weight
\cite[Proposition~4.7(1-2)]{CziszterDomokosGeroldinger2016}
\cite[\S1.4, Lemma~1.4.2]{Sturmfels2008}.

\begin{corollary}[Bases and generators for finite abelian groups and $\SO(2)$]\label{cor:finite-abelian-bases-generators}\label{cor:so2-isotypic-decomposition}\label{cor:so2-monomial-generators}
Fix bases $\set{p_{\lambda,a}}_a$ for $\HomG{X}{H_\lambda}$.
For any $d = (d_{\lambda,a})_{\lambda,a}$ with nonnegative integer entries,
write $p^d=\prod_{\lambda,a}p_{\lambda,a}^{d_{\lambda,a}}$.
Then
\begin{equation}\label{eq:finite-abelian-monomial-basis}
  \left\{p^d:
    \sum_{\lambda,a}d_{\lambda,a}=D,\quad
    \sum_{\lambda,a}\lambda d_{\lambda,a}=\nu\right\}
\end{equation}
is a basis for $\HomG{\Sym^D X}{H_\nu}$.

Moreover,
\begin{equation}\label{eq:finite-abelian-monomial-generators}
  \left\{p^d:
    d\ne0,\quad \sum_{\lambda,a}\lambda d_{\lambda,a}=\nu,\quad
    \sum_{\lambda,a}\lambda e_{\lambda,a}\ne0
    \text{ for all }0 < e < d\right\}
\end{equation}
is a minimal homogeneous generating set for
\[
  \begin{cases}
    \scR(X)\text{ as a }\dsC\text{-algebra},&\nu=0,\\
    \scM(X,H_\nu)\text{ as an }\scR(X)\text{-module},&\nu\ne0.
  \end{cases}
\]
\end{corollary}
\begin{proof}
Since $\dim H_\lambda=1$, only partitions with at most one row
contribute in Theorem~\ref{thm:isotypic-decomposition}, and
$\mathbf S_{\pi_\lambda}(H_\lambda)\cong H_{\lambda d_\lambda}$.
Thus, $\mu_\lambda=\lambda d_\lambda$, and
$\HomG{\bigotimes_\lambda H_{\mu_\lambda}}{H_\nu}$ is nonzero exactly when $\sum_\lambda \lambda d_\lambda=\nu$.
Letting $C$ and the $L_\lambda$ be scalar multiplication,
Eq.~\eqref{eq:isotypic-decomposition} simplifies to
\[
  \bigoplus_{\substack{\sum_\lambda d_\lambda=D\\
                       \sum_\lambda \lambda d_\lambda=\nu}}
  \bigotimes_\lambda\Sym^{d_\lambda}\!\left(\HomG{X}{H_\lambda}\right)
  \xrightarrow{\ \sim\ }\HomG{\Sym^D X}{H_\nu}.
\]
Up to normalization, the tableau bases on the left-hand-side evaluate under Eq.~\eqref{eq:isotypic-evaluation}
to the monomials in
Eq.~\eqref{eq:finite-abelian-monomial-basis}, with $d_\lambda=\sum_a d_{\lambda,a}$.

Each denominator in Eq.~\eqref{eq:graded-nakayama-criteria} is spanned by
products of a nonconstant invariant monomial and a monomial from the
corresponding numerator. Thus, the monomials in
Eq.~\eqref{eq:finite-abelian-monomial-generators} descend to a basis for
the quotient, and Eq.~\eqref{eq:graded-nakayama-criteria}
gives minimal generation.
\end{proof}

\subsection{\texorpdfstring{$\Orth(2)$}{O(2)}}\label{sec:examples-o2}

Let $G=\Orth(2)=\SO(2)\rtimes\langle s\rangle$, where $s$ is any fixed
reflection. The complex irreducible representations of $G$ are the trivial
representation $V_+$, the determinant representation $V_-$, and the
two-dimensional representations $V_\nu$, $\nu\in\dsN$, with
\[
  V_\pm|_{\SO(2)}\cong H_0,
  \qquad
  V_\nu|_{\SO(2)}\cong H_\nu\oplus H_{-\nu}.
\]
The reflection $s$ acts by $+1$ on $V_+$, by $-1$ on $V_-$, and by
$(z_+,z_-)\mapsto(z_-,z_+)$ on $V_\nu$.

Corollary~\ref{cor:so2-monomial-generators} and Proposition~\ref{prop:index-two-generators} analytically recover invariant and covariant generators for $\Orth(2)$ \cite[Sections~3 and~5.2]{DesmoratEtAl2023}.

\begin{corollary}[Generators for $\Orth(2)$]\label{cor:o2-generators}
Choose bases $\{p_{\lambda,a}\}_a$ for
$\Hom_{\SO(2)}(X,H_\lambda)$ so that
\[
  q_{\lambda,a}\coloneq s p_{\lambda,a}=
  \begin{cases}
    p_{-\lambda,a},&\lambda\ne0,\\
    \pm p_{0,a},&\lambda=0.
  \end{cases}
\]
Let $\Gamma_\nu$ denote the set of exponent vectors satisfying the conditions in Eq.~\eqref{eq:finite-abelian-monomial-generators}.
Then the following are homogeneous generating sets:
\begin{equation}\label{eq:o2-generators}
\begin{array}{c@{\qquad}l}
\scR_{\Orth(2)}(X)
&
\left\{p^{c+d}+q^{c+d}:
  c\in\{0\}\cup\Gamma_0,\ d\in\Gamma_0\right\}
\\[.8em]
\scM_{\Orth(2)}(X,V_+)
&
\{1\}
\\[.8em]
\scM_{\Orth(2)}(X,V_-)
&
\left\{p^c-q^c:c\in\Gamma_0\right\}
\\[.8em]
\scM_{\Orth(2)}(X,V_\nu)
&
\left\{(p^{c+d},q^{c+d}):
  c\in\{0\}\cup\Gamma_0,\ d\in\Gamma_\nu\right\}.
\end{array}
\end{equation}
The first row generates as a $\dsC$-algebra and the others as modules
over $\scR_{\Orth(2)}(X)$.
\end{corollary}
\begin{proof}
By Corollary~\ref{cor:so2-monomial-generators} and
Proposition~\ref{prop:arbitrary-output}, we have the following
homogeneous generating sets:
\[
\begin{array}{c@{\qquad}l}
\scR_{\SO(2)}(X)
&
\left\{p^c:c\in\Gamma_0\right\}
\\[.8em]
\scM_{\SO(2)}(X,V_+)
&
\{1\}
\\[.8em]
\scM_{\SO(2)}(X,V_-)
&
\{1\}
\\[.8em]
\scM_{\SO(2)}(X,V_\nu)
&
\left\{(p^d,0),(0,q^d):d\in\Gamma_\nu\right\}.
\end{array}
\]
Apply Proposition~\ref{prop:index-two-generators}.
The $V_\pm$ rows follow directly. For the other rows, use
\[
  f^-h^-=(fh)^+-f^+h^+
\]
to replace products of odd parts by even parts of products.
Taking $f=p^c$ and $h=p^d$ or $(p^d,0)$ gives the first and last
rows of Eq.~\eqref{eq:o2-generators}.
\end{proof}

\subsection{\texorpdfstring{$\SO(3)$}{SO(3)}}\label{sec:examples-so3}

Let $G=\SO(3)$. The complex irreducible representations of $G$ are the
spaces $H_\lambda$ of symmetric trace-free tensors of order
$\lambda\in\dsN_0$ on $\dsC^3$.

The Clebsch-Gordan decompositions are
\cite[Section~11.2]{FH1991}
\[
  H_\lambda\otimes H_\mu
  \cong\bigoplus_{\nu=|\lambda-\mu|}^{\lambda+\mu}H_\nu.
\]
The Clebsch-Gordan tensors can be realized as transvectants \cite[p.~88, Eq.~(5.7)]{Olver1999}
in the binary form bases for $H_\lambda$.
These are the same transvectants used in Gordan's
algorithm \cite[Section~2.2]{Olive2017}.

The Schur multiplicities are
\cite[Theorem~6.3(3) and Sections~11.1--11.2]{FH1991}
\[
  \dim\HomG{\mathbf S_\pi(H_\lambda)}{H_\nu}
  =\bigl([q^\nu]-[q^{\nu+1}]\bigr)
    s_\pi(q^{-\lambda},q^{-\lambda+1},\ldots,q^\lambda),
\]
where $s_\pi$ is the Schur polynomial and $[q^r]$ extracts the
coefficient of $q^r$.

\begin{figure}[H]
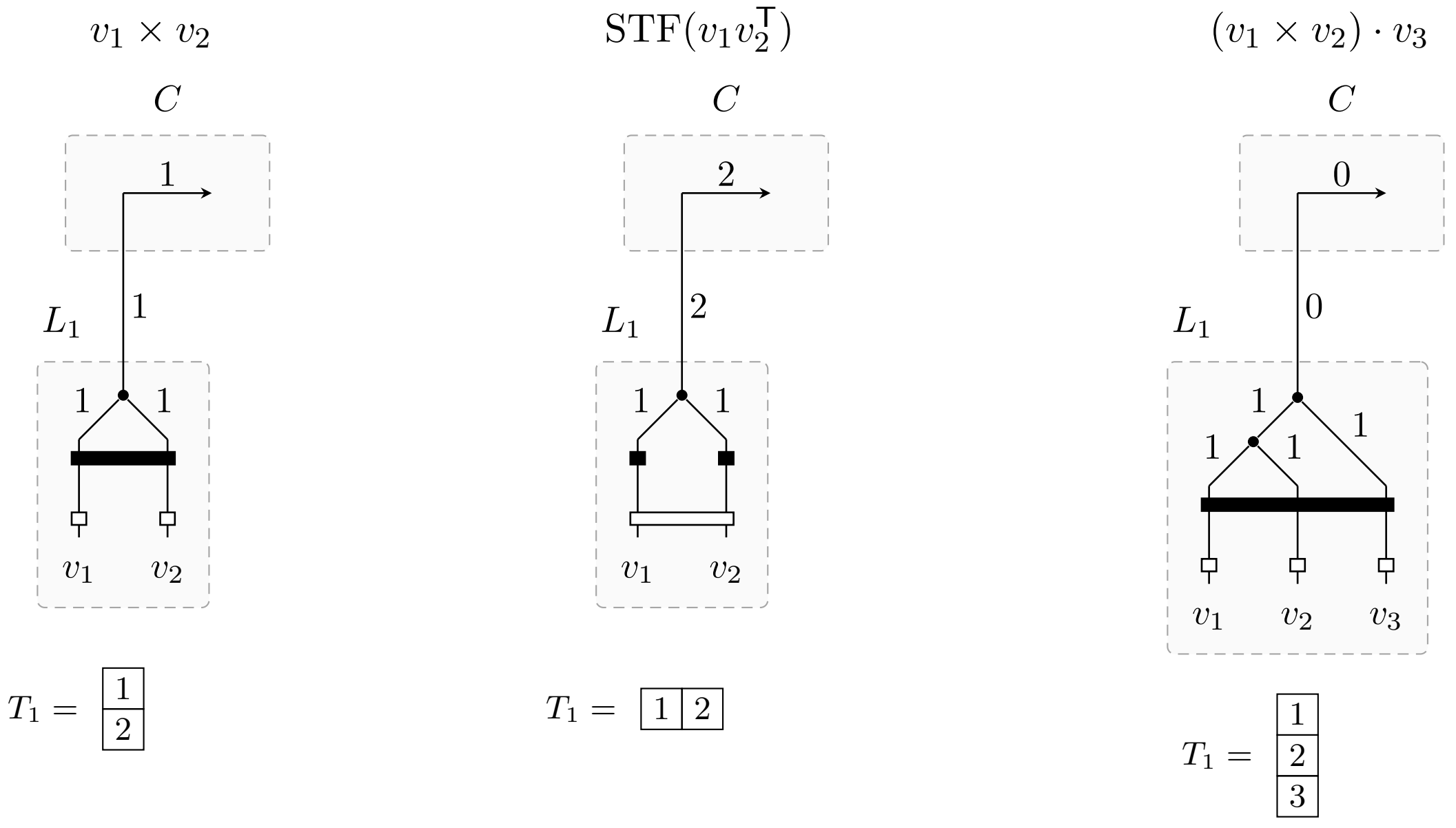

\centering
\begin{minipage}[t]{.30\linewidth}\centering
  \strut$v_1\times v_2$\par\medskip
  \begin{isotypicdiagram}[baseline=(iso-core-frame.north),iso tableau rows=3]
    \IsoLeaf[copies={v_1,v_2}]{v}{0}{1}{{1},{2}}{
      \IsoFuse{r}{s1}{s2}{1}
    }
    \IsoCore{v}{1}
  \end{isotypicdiagram}
\end{minipage}\hfill
\begin{minipage}[t]{.30\linewidth}\centering
  \strut$\operatorname{STF}(v_1v_2^{\mathsf T})$\par\medskip
  \begin{isotypicdiagram}[baseline=(iso-core-frame.north),iso tableau rows=3]
    \IsoLeaf[copies={v_1,v_2}]{v}{0}{1}{{1,2}}{
      \IsoFuse{r}{s1}{s2}{2}
    }
    \IsoCore{v}{2}
  \end{isotypicdiagram}
\end{minipage}\hfill
\begin{minipage}[t]{.38\linewidth}\centering
  \strut$(v_1\times v_2)\cdot v_3$\par\medskip
  \begin{isotypicdiagram}[baseline=(iso-core-frame.north),iso tableau rows=3]
    \IsoLeaf[copies={v_1,v_2,v_3}]{v}{0}{1}{{1},{2},{3}}{
      \IsoFuse{a}{s1}{s2}{1}
      \IsoFuse{r}{a}{s3}{0}
    }
    \IsoCore{v}{0}
  \end{isotypicdiagram}
\end{minipage}
\caption{Cross product, symmetric trace-free outer product, and scalar triple product.}
\label{fig:so3-vector-diagrams}
\label{fig:so3-determinant-action}
\end{figure}

\begin{figure}[H]
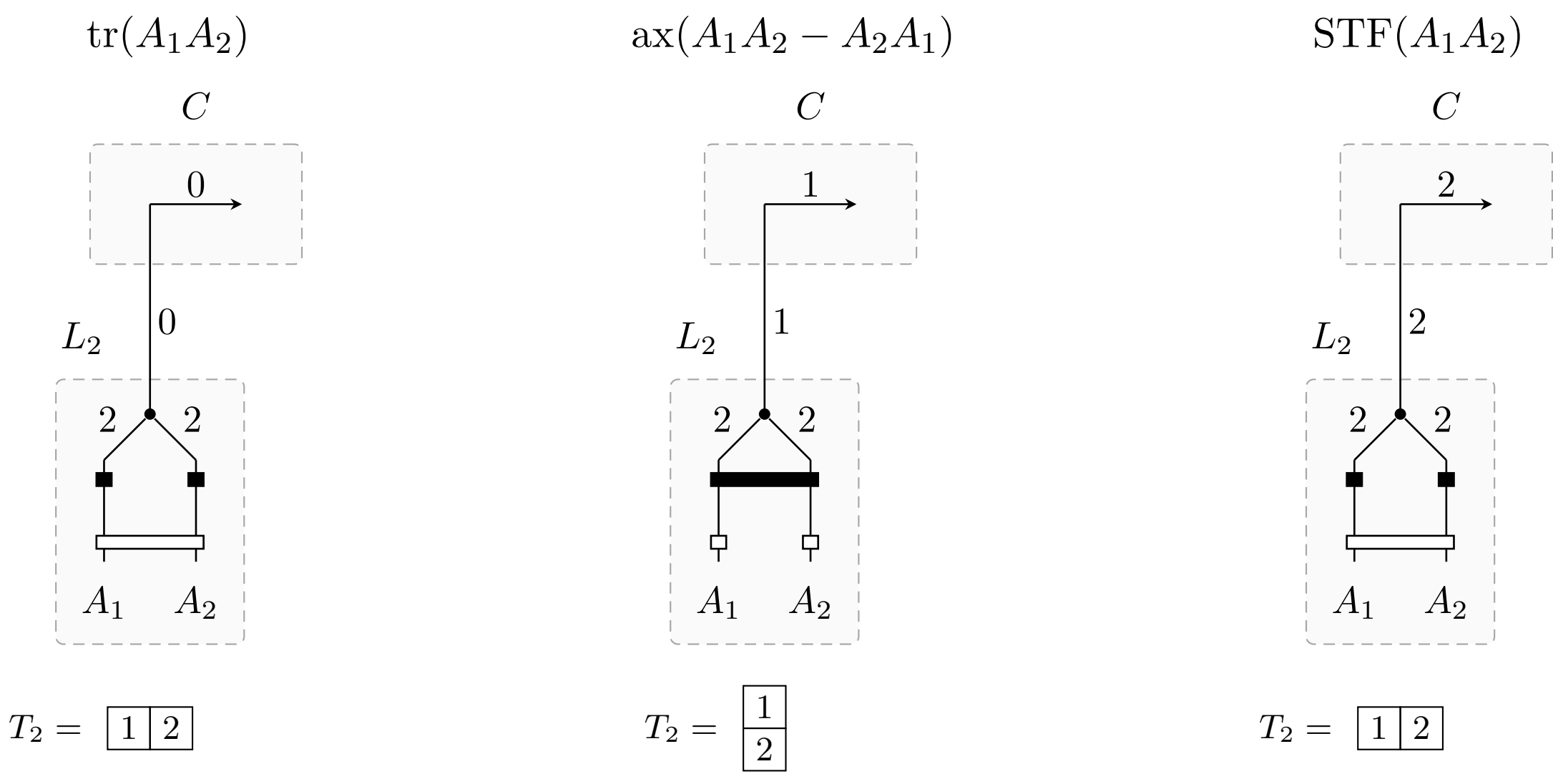

\centering
\begin{minipage}[t]{.32\linewidth}\centering
  \strut$\tr(A_1A_2)$\par\medskip
  \begin{isotypicdiagram}[baseline=(iso-core-frame.north),iso tableau rows=2]
    \IsoLeaf[copies={A_1,A_2}]{a}{0}{2}{{1,2}}{
      \IsoFuse{r}{s1}{s2}{0}
    }
    \IsoCore{a}{0}
  \end{isotypicdiagram}
\end{minipage}\hfill
\begin{minipage}[t]{.32\linewidth}\centering
  \strut$\operatorname{ax}(A_1A_2-A_2A_1)$\par\medskip
  \begin{isotypicdiagram}[baseline=(iso-core-frame.north),iso tableau rows=2]
    \IsoLeaf[copies={A_1,A_2}]{a}{0}{2}{{1},{2}}{
      \IsoFuse{r}{s1}{s2}{1}
    }
    \IsoCore{a}{1}
  \end{isotypicdiagram}
\end{minipage}\hfill
\begin{minipage}[t]{.32\linewidth}\centering
  \strut$\operatorname{STF}(A_1A_2)$\par\medskip
  \begin{isotypicdiagram}[baseline=(iso-core-frame.north),iso tableau rows=2]
    \IsoLeaf[copies={A_1,A_2}]{a}{0}{2}{{1,2}}{
      \IsoFuse{r}{s1}{s2}{2}
    }
    \IsoCore{a}{2}
  \end{isotypicdiagram}
\end{minipage}
\caption{Trace, axial commutator, and symmetric trace-free matrix product.}
\label{fig:so3-matrix-products}
\end{figure}

\begin{figure}[H]
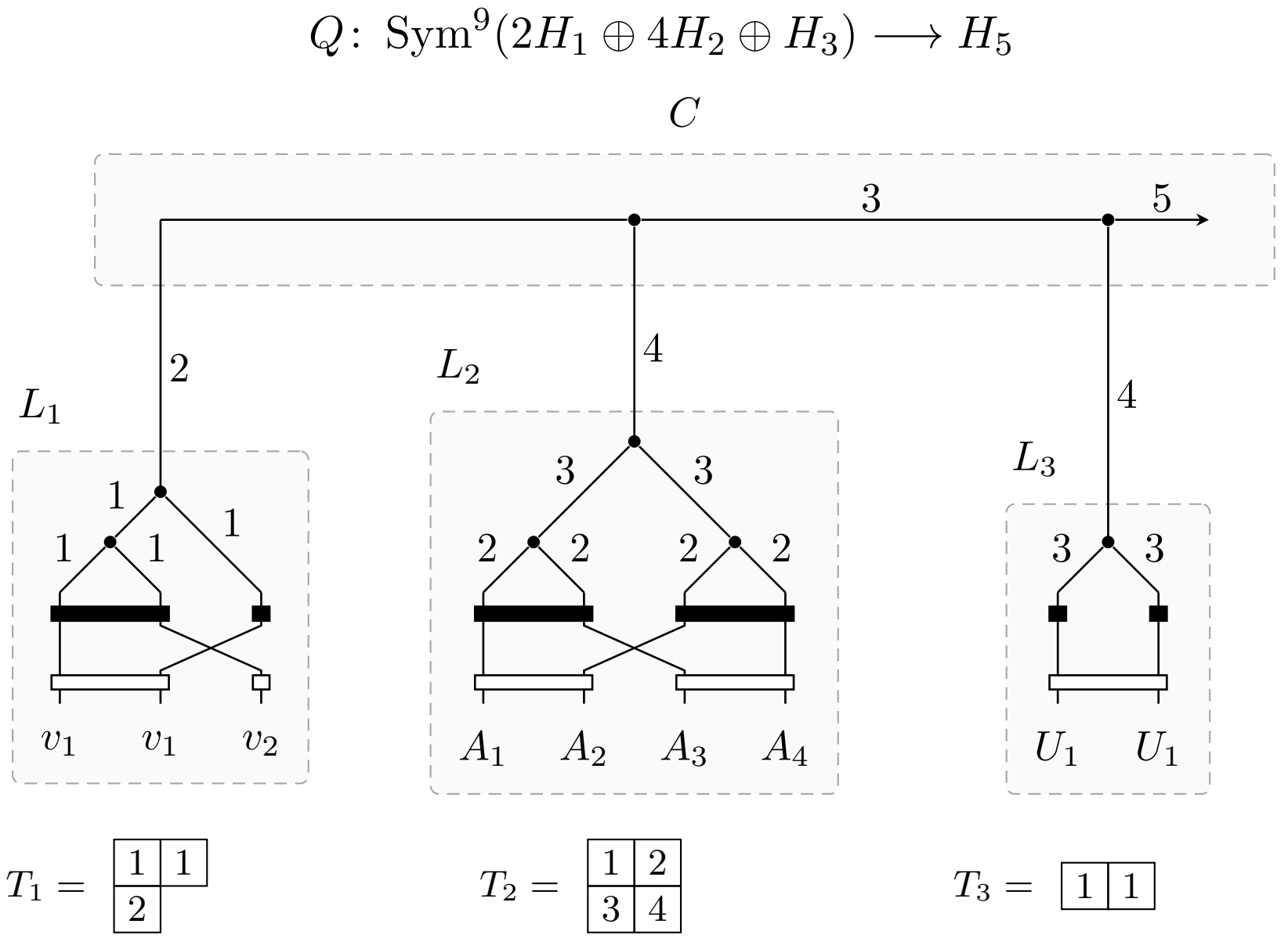

\centering
\strut$Q\colon\Sym^9(2H_1\oplus4H_2\oplus H_3)\longrightarrow H_5$\par\medskip
\begin{isotypicdiagram}
  \IsoLeaf[copies={v_1,v_2}]{v}{0}{1}{{1,1},{2}}{
    \IsoFuse{column}{s1}{s3}{1}
    \IsoFuse{r}{column}{s2}{2}
  }
  \IsoLeaf[copies={A_1,A_2,A_3,A_4}]{a}{4}{2}{{1,2},{3,4}}{
    \IsoFuse{left}{s1}{s3}{3}
    \IsoFuse{right}{s2}{s4}{3}
    \IsoFuse{r}{left}{right}{4}
  }
  \IsoLeaf[copies={U_1}]{u}{8}{3}{{1,1}}{
    \IsoFuse{r}{s1}{s2}{4}
  }
  \IsoCore{v,a,u}{3,5}
\end{isotypicdiagram}
\caption{Degree-nine covariant from $X = 2H_1\oplus4H_2\oplus H_3$ to $Y = H_5$.}
\label{fig:so3-large-diagram}
\end{figure}

Continuing Example~\ref{ex:isotypic-data-so3},
Figures~\ref{fig:so3-vector-diagrams}-\ref{fig:so3-large-diagram}
pair various invariants and covariants with their isotypic data (up to normalization). We view $v_i\in H_1$, $A_i\in H_2$, and
$U_i\in H_3$ as vectors, symmetric trace-free matrices, and symmetric trace-free 3-tensors, respectively.
We write $\operatorname{STF}$ for symmetrization and
trace removal, and define $\operatorname{ax}(K)$ for skew-symmetric $K$ by
$\operatorname{ax}(K)\times z=Kz$.

\begin{figure}[H]
\centering
\includegraphics[width=.95\linewidth]{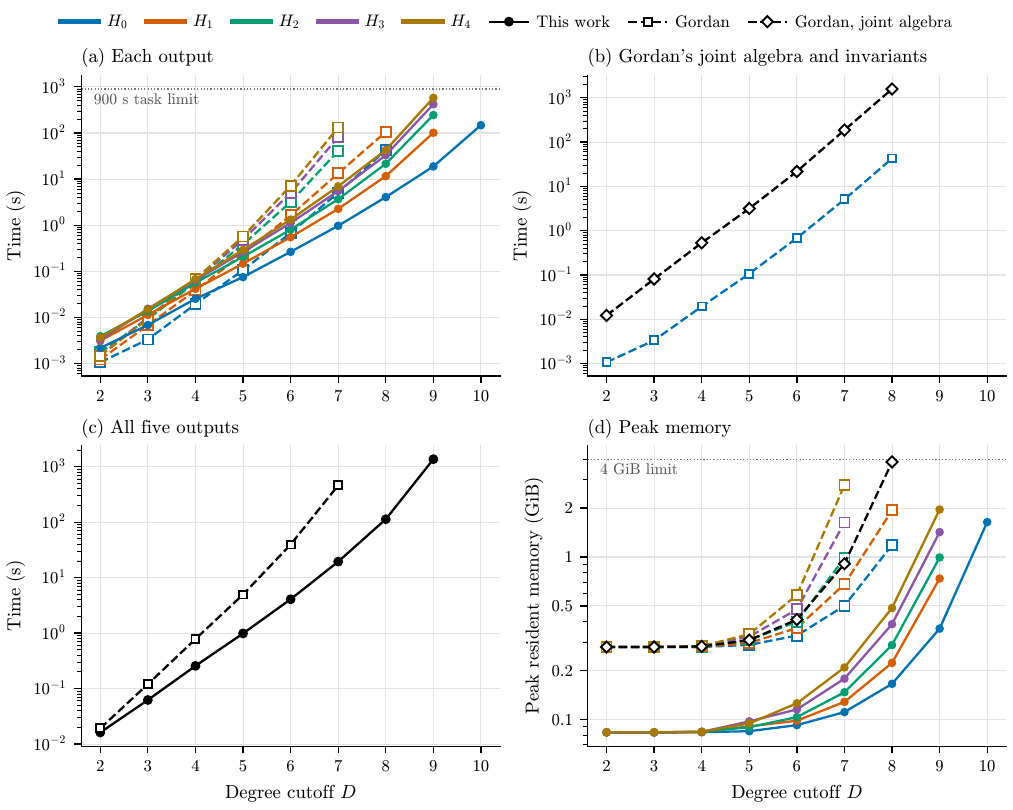}
\caption{Degree scaling for $n=1$, $X=3H_1\oplus2H_2\oplus H_3\oplus H_4$ and
$Y=H_\nu$. Colors identify $\nu$; black marks totals and the joint algebra.
Solid lines with filled markers show our method, dashed lines with open
markers Gordan's algorithm. (a)~Time per output; for Gordan, given the joint
algebra and, for $\nu>0$, the invariants. (b)~Gordan's joint algebra, summed
over its separately limited adjunction steps, and invariant extraction (the
Gordan $H_0$ curve of (a)). (c)~Total time for all five outputs. (d)~Peak
memory; for the joint algebra, the maximum over its steps. Missing points:
Gordan's $H_2$, $H_3$ and $H_4$ extractions at $D=8$ and joint algebra at
$D=9$ exceeded the memory limit, so Gordan's $D=10$ was not attempted, and
our $H_1$, $H_2$, $H_3$ and $H_4$ at $D=10$ exceeded the time limit.}
\label{fig:gordan-degree}
\end{figure}

\begin{figure}[H]
\centering
\includegraphics[width=.95\linewidth]{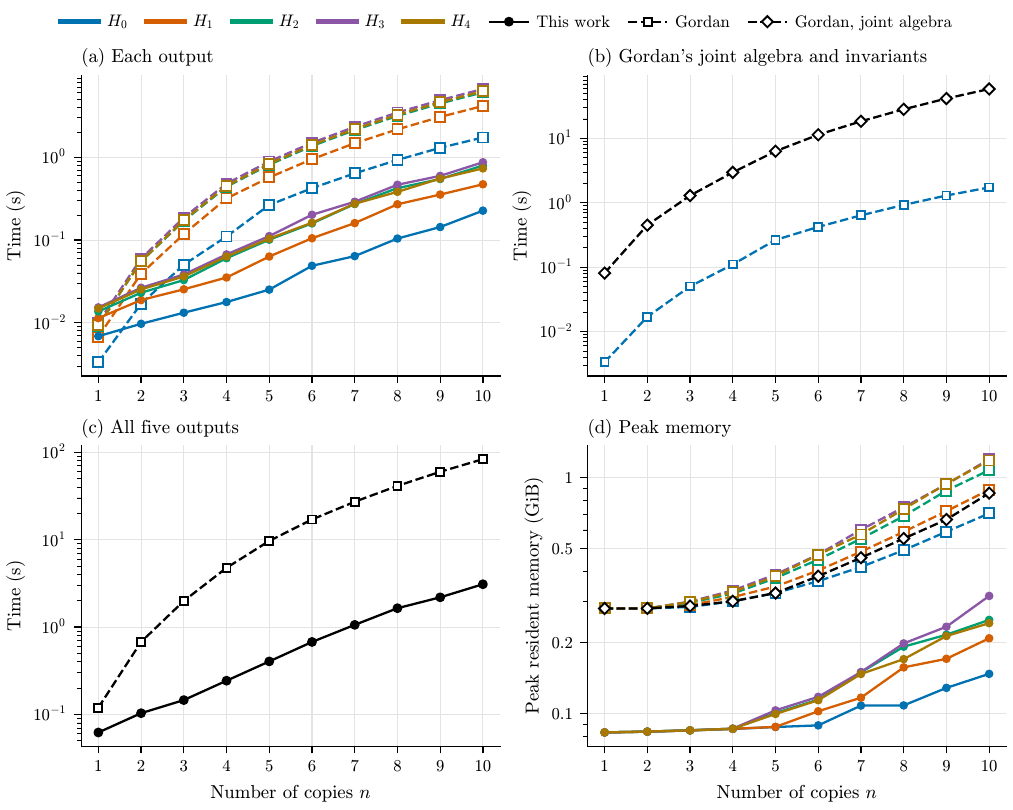}
\caption{Input-size scaling for $X_n=n(3H_1\oplus2H_2\oplus H_3\oplus H_4)$ at
$D=3$, with the conventions of Figure~\ref{fig:gordan-degree}.}
\label{fig:gordan-input}
\end{figure}

We compare our method with an optimized implementation of
Gordan's algorithm \cite{Olive2017} for

\[
X_n=n(3H_1\oplus2H_2\oplus H_3\oplus H_4),\qquad Y=H_\nu,\quad \nu=0,\ldots,4.
\]
Both methods compute the minimal homogeneous generators of degree at most $D$
of the invariant algebra ($\nu=0$) and of the covariant module ($\nu>0$). We
vary $D=2,\ldots,10$ at $n=1$ (Figure~\ref{fig:gordan-degree}) and
$n=1,\ldots,10$ at $D=3$ (Figure~\ref{fig:gordan-input}).  Each pair $(n,D)$ is
a separate computation. Generator counts agree on all 77 distinct problems
$(n,D,\nu)$ where both methods complete.
For each $(n,D)$, total times include one joint-algebra computation, one
invariant extraction and four module extractions for Gordan, and five
independent calls for our method.

All computations used one CPU thread on an Intel Core i5-14600K,
with limits of 900~s and 4~GiB per task, applied separately to each
joint algebra step. Timings exclude warm-up and checkpoint I/O.
Peak memory includes worker overhead and loaded prerequisites.

Figures~\ref{fig:gordan-degree} and~\ref{fig:gordan-input}
show speedups of  $\sim$ \!\!$25\times$ in total runtime on the largest
problems completed by both methods, with lower peak memory.
In the degree-scaling experiment, our method computes all five outputs
through $D=9$, while Gordan's algorithm sometimes exceeds the memory limit
at $D=8$.

\subsection{\texorpdfstring{$\mathrm{SU}(2)$}{SU(2)} and related groups}\label{sec:examples-su2}

For $G=\mathrm{SU}(2)$, $\mathrm{SL}(2,\dsR)$, or
$\mathrm{SL}(2,\dsC)$, consider the irreducible representations
$V_n=\Sym^n\dsC^2$, $n\in\dsN_0$.

The Clebsch-Gordan decompositions are \cite[Section~2.2]{Olive2017} \cite[Section~11.2]{FH1991}
\[
  V_a\otimes V_b\cong
  \bigoplus_{r=0}^{\min(a,b)}V_{a+b-2r}.
\]
The Clebsch-Gordan tensors can be realized as transvectants \cite[Section~2.2]{Olive2017}.

The Schur multiplicities are
\cite[Theorem~6.3(3) and Sections~11.1--11.2]{FH1991}
\[
  \dim\HomG{\mathbf S_\pi(V_n)}{V_m}
  =\bigl([q^m]-[q^{m+2}]\bigr)
    s_\pi(q^{-n},q^{-n+2},\ldots,q^n).
\]
Here again $s_\pi$ is the Schur polynomial and $[q^r]$ extracts the
coefficient of $q^r$.

\subsection{\texorpdfstring{$\Orth(3)$}{O(3)}}\label{sec:examples-o3}

Let $G=\Orth(3)=\SO(3)\times\langle s\rangle$, where $s=-I$.
The complex irreducible representations of $G$ are
$H_\lambda^\epsilon$, $\lambda\in\dsN_0$, $\epsilon\in\{+,-\}$,
with $H_\lambda^\epsilon|_{\SO(3)}\cong H_\lambda$ and $s$
acting by multiplication by $\epsilon$.

Let $X$ be a finite-dimensional complex $G$-representation.
View $X$ as an $\SO(3)$-representation and choose bases
$\set{p_{\lambda,a}}_a$ for the multiplicity spaces
$\Hom_{\SO(3)}(X,H_\lambda)$ with definite parity
$p_{\lambda,a}\circ s=\pm p_{\lambda,a}$.
With these bases, the generators $m$ returned by
Algorithm~\ref{alg:extract-independent-generators} have definite
parity $m\circ s=(-1)^{d_-}m$, where $d_-$ counts tableau entries
$p_{\lambda,a}$ of negative parity.
So Proposition~\ref{prop:index-two-generators} yields algebra and module
generators by inspecting parity.

\subsection{Point clouds in \texorpdfstring{$\Orth(n)$}{O(n)}}\label{sec:examples-on}

Let $G=\Orth(n)$. Write $H=\dsC^n$ for the standard representation.  We take $X=mH$.

Theorem~\ref{thm:isotypic-decomposition} analytically recovers the classical
generators for invariants~\cite[\S10.2]{KraftProcesi1996} and vector-valued covariants~\cite[Section~6]{BlumSmithVillar2023} by specializing
to the invariant part of the Cauchy decomposition.

\begin{corollary}[Minimal generators for point clouds]\label{cor:on-vector-generators}
The family $\{x_i\cdot x_j:1\le i\le j\le m\}$ minimally generates
$\scR(X)$ as a $\dsC$-algebra, and the family $\{x_i:1\le i\le m\}$
minimally generates $\scM(X,H)$ as an $\scR(X)$-module.
\end{corollary}
\begin{proof}
There is a single input type, so for $H_\nu=\mathbf1$, the isomorphism in
Theorem~\ref{thm:isotypic-decomposition} specializes to
\[
  \bigoplus_{\substack{\pi\vdash D\\\ell(\pi)\le\min\{n,m\}}}
    \HomG{\mathbf S_\pi(H)}{\mathbf1}
    \otimes\mathbf S_\pi\!\left(\HomG{X}{H}\right)
    \xrightarrow{\ \sim\ }\HomG{\Sym^D X}{\mathbf1},
\]
where $\ell(\pi)$ counts rows.
This is the degree-$D$ invariant part of the Cauchy decomposition of
$\Sym(X^*)$. The rest of the argument is
standard.  See, e.g., \cite[Theorem~5.2.2(1)]{GW2009},
\cite[Sections~5-6]{BlumSmithVillar2023}, and Eq.~\eqref{eq:graded-nakayama-criteria}.
\end{proof}

\subsection{\texorpdfstring{$\mathfrak S_n$}{Sn}}\label{sec:examples-sn}

Let $G$ be the symmetric group on $n$ letters. The complex irreducible representations of $G$
are the Specht modules $H_\lambda$, $\lambda\vdash n$, which admit 
Young seminormal bases over $\mathbb{Q}$~\cite[Proposition~6.1]{VershikOkounkov2005}.
Write $\chi_\lambda$ for their characters.

The Clebsch-Gordan decompositions are~\cite[Corollary~2.16]{FH1991}
\[
  H_\lambda\otimes H_\mu
  \cong\bigoplus_{\nu\vdash n}g_{\lambda\mu}^{\nu}H_\nu,
  \qquad
  g_{\lambda\mu}^{\nu}
  =\frac{1}{n!}\sum_{\sigma\in\mathfrak S_n}
    \chi_\lambda(\sigma)\chi_\mu(\sigma)\chi_\nu(\sigma),
\]
where the integers $g_{\lambda\mu}^{\nu}$ are the Kronecker coefficients
\cite[Chapter~I, \S7, pp.~115-116]{Macdonald1995}.
The intertwining equations are linear over $\mathbb{Q}$, so their solution spaces admit rational bases, and clearing denominators gives integer Clebsch–Gordan tensors. 
The same intertwining equations give the input and output multiplicity
bases when rational representation matrices are supplied.

The Schur multiplicities are
\cite[Theorem~6.3(3)]{FH1991} \cite[Theorem~2.12]{FH1991}
\[
  \dim\HomG{\mathbf S_\pi(H_\lambda)}{H_\nu}
  =\frac{1}{n!}\sum_{\sigma\in\mathfrak S_n}\chi_\nu(\sigma)
    s_\pi\bigl(\xi_{\lambda,1}(\sigma),\ldots,
      \xi_{\lambda,h_\lambda}(\sigma)\bigr),
\]
where $\xi_{\lambda,1}(\sigma),\ldots,\xi_{\lambda,h_\lambda}(\sigma)$
are the eigenvalues of $\sigma\in\mathfrak S_n$ on $H_\lambda$ and
$h_\lambda=\dim H_\lambda$.

\section{Demonstrating the learning pipeline}
\label{sec:nbody-results}

In this section we demonstrate the full pipeline of precomputing invariants
and covariants and then learning a predictive model from data. We consider
two scientific problems, and show the applicability of our framework for
modeling an end-to-end predictor as well as an equivariant block within a
larger architecture.

\subsection{Predicting the electronic Hamiltonian of water}
\label{sec:water-hamiltonian}

Given the geometry of a water molecule, we predict its Kohn--Sham Hamiltonian
in a basis of $24$ atomic orbitals~\cite{SchuttEtAl2019}. This task is a
standard benchmark for equivariant networks that predict electronic
structure~\cite{YuEtAl2023,KimEtAl2025}.

\paragraph{Our model and baseline models.}

The input is the pair $x=(b_1,b_2)$ of O--H bond vectors and the output is a
symmetric matrix on the span $V$ of the atomic orbitals, so that
\[
  G=\Orth(3)\times\mathfrak S_2,\qquad X=2H_1^-,\qquad Y=\Sym^2V,
\]
where $\mathfrak S_2$ exchanges the two hydrogen atoms and
$V\cong7H_0^+\oplus4H_1^-\oplus H_2^+$ as a representation of $\Orth(3)$.
Run through degree $10$, Algorithm~\ref{alg:extract-independent-generators}
with Proposition~\ref{prop:arbitrary-output} returns $R=3$ invariants, of
degrees $2$, $2$ and $4$, and $M=192$ covariants, of degrees at most $6$.
Our model is the single layer Eq.~\eqref{eq:polynomial-ansatz}, truncated to
total degree at most $D$ and fitted by ridge regression.

As controls, we rotate each molecule into a frame determined by its bond
bisector and its plane, and regress the Hamiltonian entries allowed by
symmetry on the bond lengths and the bond angle, by polynomial and by
Gaussian kernel ridge regression. Such a frame must be constructed by hand
for each molecule. We also compare with the released QHNet
model~\cite{YuEtAl2023} and with the published results of
QHFlow~\cite{KimEtAl2025}.

\paragraph{Experimental setup.}

The dataset contains $4999$ geometries from a molecular dynamics trajectory,
with Hamiltonians computed by density functional
theory~\cite{SchuttEtAl2019}. We use the split of the QHNet
code~\cite{YuEtAl2023}, with $500$ training, $500$ validation and $3999$ test
geometries. The validation set selects $D=21$ and the ridge penalty
$10^{-10}$. At this degree the model has $24\,748$ coefficients and
parameterizes every equivariant polynomial map $X\to Y$ of degree at most
$21$. The fit, including the validation search, takes under a minute on a
CPU. We report the mean absolute error (MAE) of the Hamiltonian entries and
of the five occupied orbital energies.

\paragraph{Results.}

Our single layer is more accurate than both networks, with roughly a
thousand times fewer parameters (Table~\ref{tab:water-hamiltonian}). This
comparison is specific to a molecule with three internal coordinates: the
networks are general-purpose architectures, and on larger molecules a single
global layer is not competitive with them.

The Gaussian kernel control matches the Hamiltonian MAE of our model on this
split and improves on it on two other random splits, where it attains $2.41$
and $2.72\,\mu E_{\mathrm h}$ against $4.04$ and $3.14\,\mu E_{\mathrm h}$
for our model. What our framework contributes is the equivariant basis, computed
from $G$, $X$ and $Y$ alone and without a molecular frame. Our model
interpolates but does not extrapolate: fitted to the $80\%$ of geometries
closest to the mean geometry, it attains $1.26\,\mu E_{\mathrm h}$ on
held-out geometries in that region and $78.80\,\mu E_{\mathrm h}$ outside it.

\begin{table}[htbp]
  \centering
  \small
  \setlength{\tabcolsep}{4pt}
  \begin{tabular}{lrrr}
    \toprule
    Model & Parameters & Hamiltonian MAE & Orbital energy MAE\\
    \midrule
    QHNet~\cite{YuEtAl2023} & 20,415,673$^{a}$ & 10.36 & 36.13\\
    QHFlow~\cite{KimEtAl2025} & 24,203,041$^{a}$ & 4.93$^{b}$ & 19.29$^{b}$\\
    \midrule
    Molecular-frame polynomial & 21,450 & 2.70 & 2.74\\
    Molecular-frame Gaussian kernel & 96,000 & 2.25 & 2.22\\
    \textbf{Ours} (Eq.~\eqref{eq:polynomial-ansatz}, $D=21$) & 24,748 & 2.20 & 1.82\\
    \bottomrule
  \end{tabular}
  \caption{Test MAE on the water Hamiltonian benchmark, in microhartree
  ($\mu E_{\mathrm h}$). Errors marked $^{b}$ are quoted
  from~\cite{KimEtAl2025} and use $3900$ test geometries; we evaluated all
  other models on the same $3999$ test geometries. Parameter counts marked
  $^{a}$ are obtained by instantiating the authors' released code in its
  water configuration.}
  \label{tab:water-hamiltonian}
\end{table}

\subsection{Learning charged particle dynamics}
\label{sec:nbody}

Five particles with charges $\pm1$ move under Coulomb forces in
$\dsR^3$~\cite{KipfEtAl2018}. Given their initial positions $q_i$, velocities
$v_i$ and charges, we predict their positions after $1000$ simulator steps.
This task is the standard benchmark of EGNN~\cite{SatorrasHoogeboomWelling2021}
and of the models that improve on it.

\paragraph{Our models and baseline models.}
\label{sec:nbody-model}

EGNN moves each particle by a sum, over the other particles, of $u_{ij}=q_i-q_j$
times a learned function of the two particles' features and
$\lVert u_{ij}\rVert^2$. On each pair, this is Eq.~\eqref{eq:neural-ansatz} with the
single covariant $u_{ij}$; the velocities do not enter as vector
covariants in this pairwise term.

We give each pair the three vectors $u_{ij}$, $v_i$ and $v_j$, so that
\[
  G=\Orth(3),\qquad X=3H_1^-,\qquad Y=H_1^-.
\]
Algorithm~\ref{alg:extract-independent-generators} returns, up to a linear
change of basis, the six inner products $r=(r_1,\dots,r_6)$ of these vectors
as invariants and $(m_1,m_2,m_3)=(u_{ij},v_i,v_j)$ as covariants.
According to Eq.~\eqref{eq:neural-ansatz}, our block replaces EGNN's term by
\[
  \scN_{\theta,1}(s_{ij},r)\,u_{ij}
  +\scN_{\theta,2}(s_{ij},r)\,v_i
  +\scN_{\theta,3}(s_{ij},r)\,v_j,
\]
where $s_{ij}$ collects the scalar node and edge features as in Example~\ref{ex:application-egnn}. The rest of EGNN is unchanged. This is
the layer of GMN~\cite[Eq.~(11)]{HuangEtAl2022GMN} and
EGHN~\cite{HanEtAl2022EGHN}, here obtained automatically.

We compare with EGNN at the same widths and depths, trained by us, and with
published results.

\paragraph{Experimental setup.}
\label{sec:nbody-protocol}

We follow EGNN's data, split and training protocol, with $3000$ training
systems, $10\,000$ epochs, and testing at the epoch with the lowest
validation error~\cite{SatorrasHoogeboomWelling2021}. We vary the width $w$
of the hidden layers and the number of blocks $L$, and train each
configuration with three seeds on one CPU thread. Published results are
quoted from the cited papers; their protocols differ slightly.

\paragraph{Results.}

Our block is more accurate than EGNN's at every width and depth and on every
seed, at about the same cost per block (Table~\ref{tab:nbody-configs}). Our
smallest model is more accurate than every EGNN we trained, and one of our
blocks is more accurate than four of EGNN's of the same width.

Our models do not overtake recent architectures (Table~\ref{tab:nbody}),
which redesign the whole network rather than one term. Since every particle
interacts with every other, this benchmark tests what one block can express,
not how far information travels; on sparse graphs, a richer block does not
replace message passing~\cite{JoshiEtAl2023}.

\begin{table}[htbp]
  \centering
  \small
  \setlength{\tabcolsep}{4pt}
  \begin{tabular}{lrrrrrr}
    \toprule
    Block & $w$ & $L$ & Parameters & Test MSE ($\times10^{-3}$) & Predict (ms) & Memory (MiB)\\
    \midrule
    EGNN & 16 & 4 & 8,932 & 6.66 $\pm$ 0.15 & 1.26 & 5.3\\
    EGNN & 32 & 4 & 34,244 & 7.18 $\pm$ 0.17 & 2.01 & 10.1\\
    EGNN & 64 & 1 & 33,601 & 41.92 $\pm$ 0.03 & 1.21 & 5.0\\
    EGNN & 64 & 2 & 67,074 & 7.92 $\pm$ 0.06 & 2.33 & 9.9\\
    EGNN & 64 & 4 & 134,020 & 6.46 $\pm$ 0.19 & 4.60 & 19.6\\
    \midrule
    Ours & 16 & 2 & 4,706 & 5.15 $\pm$ 0.51 & 0.97 & 2.9\\
    Ours & 16 & 4 & 9,380 & 5.27 $\pm$ 0.62 & 1.82 & 5.7\\
    Ours & 32 & 1 & 8,833 & 6.59 $\pm$ 0.19 & 0.73 & 2.7\\
    Ours & 32 & 2 & 17,602 & 5.29 $\pm$ 0.21 & 1.37 & 5.3\\
    Ours & 32 & 4 & 35,140 & 5.02 $\pm$ 0.30 & 2.62 & 10.4\\
    Ours & 32 & 6 & 52,678 & 4.89 $\pm$ 0.17 & 3.86 & 15.6\\
    Ours & 64 & 1 & 34,049 & 6.19 $\pm$ 0.13 & 1.39 & 5.1\\
    Ours & 64 & 2 & 67,970 & 5.84 $\pm$ 0.34 & 2.66 & 10.1\\
    Ours & 64 & 4 & 135,812 & 4.80 $\pm$ 0.19 & 5.18 & 20.0\\
    \bottomrule
  \end{tabular}
  \caption{Models we trained under EGNN's protocol. Rows marked \emph{Ours} use the
  three-covariant block. Test MSE is the mean $\pm$
  sample standard deviation over three seeds. Prediction time is the median
  over $15$ batches of $100$ systems on one CPU thread; memory is the autograd
  storage for one training batch.}
  \label{tab:nbody-configs}
\end{table}

\begin{table}[htbp]
  \centering
  \small
  \setlength{\tabcolsep}{4pt}
  \begin{tabular}{lrrl}
    \toprule
    Model & Test MSE ($\times10^{-3}$) & Parameters & Source\\
    \midrule
    TFN~\cite{ThomasEtAl2018} & 15.5 & 127,024$^{a}$ & \cite{SatorrasHoogeboomWelling2021}\\
    Radial Field & 10.4 & 2,052$^{a}$ & \cite{SatorrasHoogeboomWelling2021}\\
    EGNN~\cite{SatorrasHoogeboomWelling2021} & 7.1 & 134,020 & \cite{SatorrasHoogeboomWelling2021}\\
    ClofNet~\cite{DuEtAl2022ClofNet} & 6.5 & 160,964$^{a}$ & \cite{XuEtAl2024EGNO}\\
    FA-GNN~\cite{PunyEtAl2022FA} & 5.7 $\pm$ 0.2 & 92,103$^{a}$ & \cite{LevyEtAl2023MCEGNN}\\
    \textbf{Ours} ($w=16$, $L=2$) & \textbf{5.15 $\pm$ 0.51} & 4,706 & this work\\
    \textbf{Ours} ($w=32$, $L=4$) & \textbf{5.02 $\pm$ 0.30} & 35,140 & this work\\
    \textbf{Ours} ($w=64$, $L=4$) & \textbf{4.80 $\pm$ 0.19} & 135,812 & this work\\
    SEGNN~\cite{BrandstetterEtAl2022SEGNN} & 4.3 $\pm$ 0.15 & 147,424$^{b}$ & \cite{BrandstetterEtAl2022SEGNN}\\
    CN-GNN~\cite{KabaEtAl2023CN} & 4.3 $\pm$ 0.1 & 109,063$^{a}$ & \cite{LevyEtAl2023MCEGNN}\\
    PONITA~\cite{BekkersEtAl2024PONITA} & 4.3 $\pm$ 0.1 & 1,058,181$^{a}$ & \cite{BekkersEtAl2024PONITA}\\
    MC-EGNN-2~\cite{LevyEtAl2023MCEGNN} & 4.1 $\pm$ 0.6 & 135,244 & \cite{LevyEtAl2023MCEGNN}\\
    CGENN~\cite{RuheBrandstetterForre2023} & 3.9 $\pm$ 0.1 & 134,625$^{a}$ & \cite{RuheBrandstetterForre2023}\\
    CG-EGNN-1-2~\cite{TranEtAl2025CGEGNN} & 3.5 $\pm$ 0.1 & -- & \cite{TranEtAl2025CGEGNN}\\
    \bottomrule
  \end{tabular}
  \caption{Test MSE on the charged $N$-body benchmark. Parameter counts are
  stated in the cited papers unless marked: $^{a}$ derived from the paper's
  settings and the authors' code; $^{b}$ reported by Manolache et
  al.\ (arXiv:2505.13631). ``--'' marks counts that are neither stated nor
  derivable.}
  \label{tab:nbody}
\end{table}

\section{Conclusion}\label{sec:conclusion}

We developed a framework that unifies constructions in the equivariant machine learning literature, from sparse vector recovery and equivariant networks to turbulence modeling and interatomic potentials (Section~\ref{sec:applications}). The approach is to build equivariance into the features rather than into the architecture, so that the coefficient functions can be any regressor, from linear least squares to a deep network.

Our \href{https://github.com/MichaelLiu2024/Equivariant-Polynomial-Tensor-Functions}{Python package} makes this framework easy to use.
It computes the generators automatically and accepts any PyTorch module or any estimator with fit and predict methods as the coefficient model.
Many of our experiments can be reproduced with simple calls to the package.

We developed Algorithm \ref{alg:extract-independent-generators} for computing minimal homogeneous generators for the algebra of invariant polynomials and the module of equivariant polynomials up to given degrees.
Our algorithm works for any group and any finite-dimensional semisimple representations, provided that the computational requirements in Assumption \ref{ass:computational} are satisfied. This is true for many groups (see Section \ref{sec:examples-algebra}) arising in applications. We demonstrated the computational efficiency of the algorithm in Figures~\ref{fig:gordan-degree} and~\ref{fig:gordan-input}.

Several questions remain for future work.
We encode candidate algebra and module generators combinatorially via their isotypic data; might redundancies among these generators be detected directly from this data?
How should we subselect features for a particular learning task when minimal generating sets are too large?
Can we build on Proposition~\ref{prop:degree-truncation} to understand how minimality of the generating sets affects both approximation and learning efficiency when the coefficient models have limited expressivity?
Finally, what tradeoffs between computational cost and prediction accuracy occur when approximate invariant and equivariant features are used?

\section*{Acknowledgments}
J.K. is grateful to Sahil Kommalapati and Robert Moser for related collaborations.  M.L.L. is grateful to Sahil Kommalapati for helpful conversations.  This research was supported in part by NSF DMS 2309782, NSF DMS 2436499, NSF CISE-IIS 2312746,
DE SC0025312, and the Sloan Foundation.

\section*{AI Use Disclosure}

Key ideas in this work are the authors' own: considering the general ansatz Eq.~\eqref{eq:neural-ansatz}, using isotypic decompositions in Theorem~\ref{thm:isotypic-decomposition}, using random streaming to accelerate the vector space basis computations, using Nakayama's lemma to extract minimal generators, and realizing that Eq.~\eqref{eq:neural-ansatz} could apply to end-to-end predictive models as well as constitute equivariant blocks in layers in larger architectures.
GPT-6 Astra was used to review the literature, write and debug code, and draft preliminary portions of the manuscript which we scrutinized and revised thoroughly.
Each line of AI-assisted work has been examined and revised by the authors, and we take responsibility for the final content of this work.

\bibliographystyle{plain}
\bibliography{references}

\appendix

\section{Tableau, tensor train, and tensor tree bases}
\label{sec:appendix-preliminaries}
\label{sec:appendix-implementation}

We construct bases for the three tensor factors in
Eq.~\eqref{eq:isotypic-decomposition}, as required in the final step of
Algorithm~\ref{alg:stream-vector-space-basis}.
Then, we explain how to evaluate the resulting covariants in Eq.~\eqref{eq:isotypic-evaluation}.

\subsection{Tableau bases for Schur powers}
\label{sec:appendix-schur-evaluation}

Let $\pi\vdash d$ be an integer partition. A \emph{Young tableau} of shape $\pi$ is a filling
by positive integers of the left-justified diagram with $\pi_i$ boxes
in row $i$. A Young tableau is \emph{semistandard} if its entries weakly
increase along rows and strictly increase down columns.

Let $R_\pi,C_\pi\le\mathfrak S_d$ be the row and column stabilizers
of the tableau of shape $\pi$ obtained by filling its boxes
with $1,\ldots,d$ in row-major order. The \emph{Young symmetrizer} is
$e_\pi=c_\pi r_\pi\in\dsC[\mathfrak S_d]$, where
\[
  c_\pi=\sum_{\sigma\in C_\pi}
                                  \operatorname{sgn}(\sigma)\sigma,
  \qquad
  r_\pi=\sum_{\sigma\in R_\pi}\sigma.
\]
For any finite-dimensional complex vector space $V$, the associated Schur functor is
$\mathbf S_\pi(V)=e_\pi(V^{\otimes d})$.

Given a basis $\{v_a\}_a$ for $V$, we identify a tableau
$T$ having entries in $\{1,\ldots,\dim V\}$ with the tensor
\[
  T=e_\pi(v_{a_1}\otimes\cdots\otimes v_{a_d})\in\mathbf S_\pi(V),
\]
where the $a_k$ are read from the boxes of $T$ in row-major order.

\begin{lemma}[Semistandard Young tableau basis]\label{lem:ssyt-basis}
The tensors $T\in\mathbf S_\pi(V)$ associated with the
semistandard tableaux of shape $\pi$ having entries in
$\{1,\ldots,\dim V\}$ form a basis for $\mathbf S_\pi(V)$.
\end{lemma}
\begin{proof}
See \cite[Exercise~6.14 and Problem~6.15]{FH1991}.
\end{proof}

To define $T(x)$ in Eq.~\eqref{eq:isotypic-evaluation}, choose a basis
$\{p_{\lambda,a}\}_a$ for $V=\HomG{X}{H_\lambda}$. Let
\begin{equation}\label{eq:appendix-tableau-input}
  T(x)\coloneq e_\pi\!
    \left(p_{\lambda,a_1}(x)\otimes\cdots\otimes p_{\lambda,a_d}(x)\right)
  \in\mathbf S_\pi(H_\lambda)
\end{equation}
and extend linearly to all $T\in\mathbf S_\pi(V)$.
This yields a homogeneous degree-$d$ $G$-equivariant polynomial $x\mapsto T(x)$,
which is independent of the chosen basis since evaluation commutes with $e_\pi$.

\subsection{Tensor-train bases for tensor product multiplicity spaces}\label{sec:prelim-tensor-trains}

Fix Clebsch-Gordan bases for
$\HomG{H_\alpha\otimes H_\beta}{H_\gamma}$.
Here, a \emph{tensor train} \cite[Section~2]{Oseledets2011} is a chain where vertices specify Clebsch-Gordan maps and edges specify irreducible representations \cite[Eq.~(15)]{SinghPfeiferVidal2010}. A tensor train represents a tensor by contracting along shared edges.

\begin{figure}[H]
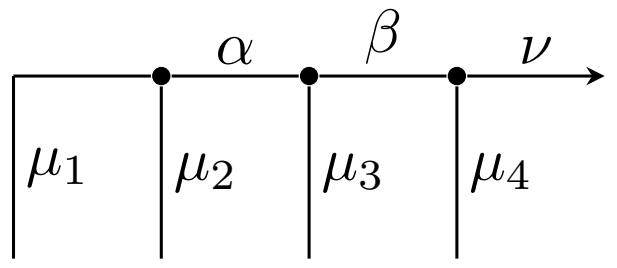

  \centering
  \scalebox{1.15}{%
  \begin{tensordiagram}
    \TensorInput{v1}{0}{}{\mu_1}
    \TensorInput{v2}{1}{}{\mu_2}
    \TensorInput{v3}{2}{}{\mu_3}
    \TensorInput{v4}{3}{}{\mu_4}
    \TensorTrain{v1,v2,v3,v4}{\alpha,\beta,\nu}
  \end{tensordiagram}%
  }
  \caption{A tensor train from
  $H_{\mu_1}\otimes H_{\mu_2}\otimes H_{\mu_3}\otimes H_{\mu_4}$
  to $H_\nu$, read from left to right. For example,
  $H_\alpha$ may be any irreducible constituent of
  $H_{\mu_1}\otimes H_{\mu_2}$, and the first vertex may be any element of the chosen basis for $\HomG{H_{\mu_1}\otimes H_{\mu_2}}{H_\alpha}$.}
  \label{fig:tensor-trains-and-trees}
\end{figure}

\begin{proposition}[Tensor train basis]\label{prop:cg-tensor-train-basis}
The tensor trains obtained by varying all admissible edges and all admissible vertices yield a basis for $\HomG{H_{\mu_1}\otimes\cdots\otimes H_{\mu_t}}{H_\nu}$.
\end{proposition}
\begin{proof}
Repeated isotypic decomposition identifies the above Hom space
with a direct sum over all admissible edges, where each direct summand is a tensor
product over all vertices of multiplicity spaces of the form $\HomG{H_\alpha\otimes H_\beta}{H_\gamma}$ \cite[Section~4.9]{EGNO2015}. The direct sum of product bases corresponds to the tensor trains in the statement.
\end{proof}

\subsection{Tensor tree bases for Schur power multiplicity spaces}
\label{sec:appendix-subroutines}
\label{sec:appendix-leaf-bases}

We obtain a basis for $\HomG{\mathbf S_\pi(H_\lambda)}{H_\mu}$ by considering elements of the form $L\circ e_\pi$, where $L$ is a \emph{tensor tree}
\cite{ShiDuanVidal2006} mapping $H_\lambda^{\otimes d}$ to $H_\mu$. In our implementation, a tensor train
couples the copies of $H_\lambda$ associated with each column of $\pi$ to a single
irreducible output. Then, another tensor train couples these column
outputs to $H_\mu$.

\begin{figure}[H]
  \centering
  \begin{isotypicdiagram}
    \IsoLeaf[enclosure=false,copies={1,2,3,4,5}]
      {schur}{0}{\lambda}{{1,2},{3,4},{5}}{
      \IsoFuse{column1a}{s1}{s3}{\alpha}
      \IsoFuse{column1}{column1a}{s5}{\eta_1}
      \IsoFuse{column2}{s2}{s4}{\eta_2}
      \IsoFuse{root}{column1}{column2}{\mu}
    }
    \IsoLeafOutput{schur}
  \end{isotypicdiagram}
  \caption{A map $L\circ e_\pi$ for $\pi=(2,2,1)$. 
  $L$ is represented by the tensor tree in the top half of the diagram. 
   The bottom half of the diagram represents the Young symmetrizer
$e_\pi$ via row symmetrization and column antisymmetrization
\cite{Cvitanovic2008}. 
  The two column trains have outputs $H_{\eta_1}$ and $H_{\eta_2}$,
  which are coupled to $H_\mu$.}
  \label{fig:schur-leaf-expansion}
\end{figure}

With the tree shape fixed by the partition $\pi$, varying all admissible edge labels and vertex
maps gives a basis for $\HomG{H_\lambda^{\otimes d}}{H_\mu}$ \cite[Eq.~(15)]{SinghPfeiferVidal2010}.
Precomposing these maps with $e_\pi$ yields a spanning set for $\HomG{\mathbf S_\pi(H_\lambda)}{H_\mu}$, but this may introduce linear dependencies.
We remove these dependencies in two stages, using the factorization
\[
  L\circ e_\pi=(L\circ c_\pi)\circ r_\pi.
\]

First, we incorporate the column antisymmetrizers into the column trains.
If $h_1,\ldots,h_s$ are the column heights of $\pi$, then
\[
  c_\pi(H_\lambda^{\otimes d})
  \cong\bigotimes_{j=1}^s\Lambda^{h_j}(H_\lambda).
\]
For each column $j$ and output $\eta_j$, the column trains restrict to a
spanning set for $\HomG{\Lambda^{h_j}(H_\lambda)}{H_{\eta_j}}$ by semisimplicity.
Evaluate the column trains on $w_1\wedge\cdots\wedge w_{h_j}$ with independently sampled
$w_i\in H_\lambda$.
Stack the sampled outputs into one matrix column per train and select
independent columns until reaching $\dim\HomG{\Lambda^{h_j}(H_\lambda)}{H_{\eta_j}}$, which is known by Assumption~\ref{ass:computational}.
Couple the selected maps using the tensor train bases of
$\HomG{H_{\eta_1}\otimes\cdots\otimes H_{\eta_s}}{H_\mu}$,
varying all $\eta_j$. By Proposition~\ref{prop:cg-tensor-train-basis},
the resulting maps $L\circ c_\pi$ form a basis for
$\HomG{c_\pi(H_\lambda^{\otimes d})}{H_\mu}$.

Second, precompose with $r_\pi$ to restrict further to
$\mathbf S_\pi(H_\lambda)$. By semisimplicity, the maps $L\circ e_\pi$
still span $\HomG{\mathbf S_\pi(H_\lambda)}{H_\mu}$.
Sample one vector $v_i\in H_\lambda$ for each row $i=1,\ldots,\ell$
and evaluate $L$ on
\begin{equation}\label{eq:schur-row-input}
  c_\pi\!\left(\bigotimes_{i=1}^\ell v_i^{\otimes\pi_i}\right).
\end{equation}
Apply the same rank selection to retain
$\dim\HomG{\mathbf S_\pi(H_\lambda)}{H_\mu}$ independent maps.

\begin{proposition}[Tensor tree basis for Schur power multiplicity spaces]
\label{prop:general-schur-tree-spanning}
If the above procedure succeeds, the resulting trees $\set L$ restrict to a basis $\set{L\circ e_\pi}$ for
$\HomG{\mathbf S_\pi(H_\lambda)}{H_\mu}$.
\end{proposition}

\subsection{Evaluation from isotypic data}
\label{sec:appendix-evaluation}
\label{sec:appendix-polarization-proof}

\begin{proposition}[Row polarization]
\label{prop:waring-grid-evaluation}
Let $\alpha_1,\ldots,\alpha_t$ be positive integers, let $k=\sum_a\alpha_a$,
and let $V$ be a vector space over a field of characteristic zero or $p>k$.
For any $v_1,\ldots,v_t\in V$,
\begin{equation}\label{eq:waring-row-evaluation}
  \Sym\!\left(v_1^{\otimes\alpha_1}\otimes\cdots\otimes
              v_t^{\otimes\alpha_t}\right)
  =\sum_{0\le j_a\le\alpha_a}\left(\prod_a(-1)^{\alpha_a-j_a}\binom{\alpha_a}{j_a}\right)
    \left(\sum_{a=1}^t j_av_a\right)^{\otimes k},
\end{equation}
where $\Sym$ sums over all $k!$ permutations.
\end{proposition}
\begin{proof}
This is the polarization identity \cite[Eq.~(7)]{Thomas2014} applied to
symmetric tensor products, with repeated arguments grouped together.
\end{proof}

We evaluate the covariant $Q$ from its isotypic data as follows. For each $\lambda$, compute the projections
$p_{\lambda,a}(x)$ in Eq.~\eqref{eq:appendix-tableau-input}. Apply Eq.~\eqref{eq:waring-row-evaluation} to each row in $e_{\pi_\lambda}$ to express
$T_\lambda(x)$ as a linear combination of tensors of the form in Eq.~\eqref{eq:schur-row-input}. Evaluate $L_\lambda$ on these tensors and form the linear combination. Finally, apply
$C$ to the resulting outputs. Our implementation takes advantage of computation sharing and other optimizations enabled by Proposition \ref{prop:waring-grid-evaluation}.

\section{Proofs}
\label{sec:appendix-proofs}

\subsection{Proof of Proposition~\ref{prop:degree-truncation}}
\label{sec:proof-expressivity}

For (i), let $F$ be in the uniform closure of models in Eq.~\eqref{eq:polynomial-ansatz}.  The approximation hypothesis on $\mathcal{N}_{\theta}$ and fact that the $m_i$ are bounded on $K$ imply that every $F$ uniformly approximated by models in  Eq.~\eqref{eq:polynomial-ansatz}
can be uniformly approximated by models in Eq.~\eqref{eq:neural-ansatz}.
Next, consider $F$ such that $\sum_{i=1}^M \mathcal{N}_{\theta_n, i}(r(x)) m_i(x) \rightarrow F$ uniformly on $K$ as $n \rightarrow \infty$ where $r = (r_1, \ldots, r_R)|_K : K \rightarrow \mathbb{R}^R$.
For each $x \in K$, this restricts to uniform convergence of maps restricted to the fiber $r^{-1}(r(x))$ in $K$. %
By closedness of finite-dimensional subspaces, it follows that $F|_{r^{-1}(r(x))}$ lies in the $\mathbb{R}$-span of $m_1|_{r^{-1}(r(x))}, \ldots, m_M|_{r^{-1}(r(x))}$.  
So there exists $a : \mathbb{R}^R \rightarrow \mathbb{R}^M$ such that $F(x) = \sum_{i=1}^M a_i(r(x))m_i(x)$ for all $x \in K$.
Thirdly, assume $F$ is such that $a$ exists as announced. 
Separately on each fiber of $r$, the expression for $F$ agrees with a model in Eq.~\eqref{eq:polynomial-ansatz}. 
By \cite[Corollary~2.9(b)]{Chen2019}, applied on $K$ with $[0,1]$-valued polynomial multipliers in $r$, this implies
uniform approximation by models in Eq.~\eqref{eq:polynomial-ansatz}.

For (ii), the assumptions that $r_j$ and $m_i$ are generators  make
Eq.~\eqref{eq:polynomial-ansatz} the module of all equivariant
polynomial maps. 
For compact groups $G$, the uniform closure on compact subsets $K \subset X$ is all equivariant continuous maps, see the proof of \cite[Proposition~2.4]{Yarotsky2022}. 
With (i), this gives the conclusion about Eq.~\eqref{eq:neural-ansatz}. \qed

\subsection{Proof of Theorem~\ref{thm:isotypic-decomposition}}
\label{sec:proof-main-decomposition}

Recall that the isotypic decomposition of $X$ is
\[
  X\cong\bigoplus_\lambda H_\lambda\otimes\HomG{H_\lambda}{X}.
\]
Take the symmetric power on both sides and distribute the symmetric power over the direct sum \cite[Chapter~III, \S6.6, Proposition~9]{Bourbaki1989} to obtain
\[
  \Sym^D X\cong
  \Sym^D\!\left(\bigoplus_\lambda H_\lambda\otimes\HomG{H_\lambda}{X}\right)
  \cong\bigoplus_{(d_\lambda)_\lambda}
  \bigotimes_\lambda\Sym^{d_\lambda}\!\left(H_\lambda\otimes\HomG{H_\lambda}{X}\right).
\]
Apply the Cauchy decomposition \cite[Corollary~2.3.3]{Weyman2003} on the right-hand-side and rearrange the tensor product to obtain
\[
  \Sym^D X\cong
  \bigoplus_{(d_\lambda)_\lambda}\bigotimes_\lambda
  \bigoplus_{\pi_\lambda}
  \mathbf S_{\pi_\lambda}(H_\lambda)
    \otimes\mathbf S_{\pi_\lambda}\!\left(\HomG{H_\lambda}{X}\right)
  \cong
  \bigoplus_{(d_\lambda)_\lambda}\bigoplus_{(\pi_\lambda)_\lambda}
  \bigotimes_\lambda\mathbf S_{\pi_\lambda}(H_\lambda)
    \otimes\bigotimes_\lambda\mathbf S_{\pi_\lambda}\!\left(\HomG{H_\lambda}{X}\right).
\]
Substitute the isotypic decomposition
\[
  \mathbf S_{\pi_\lambda}(H_\lambda)
  \cong\bigoplus_{\mu_\lambda}H_{\mu_\lambda}
  \otimes\HomG{H_{\mu_\lambda}}{\mathbf S_{\pi_\lambda}(H_\lambda)}
\]
of each Schur power on the right-hand-side and rearrange the tensor product to obtain
\[
  \Sym^D X\cong
  \bigoplus_{(d_\lambda)_\lambda}\bigoplus_{(\pi_\lambda)_\lambda}
  \bigoplus_{(\mu_\lambda)_\lambda}
  \bigotimes_\lambda H_{\mu_\lambda}
  \otimes\bigotimes_\lambda
    \HomG{H_{\mu_\lambda}}{\mathbf S_{\pi_\lambda}(H_\lambda)}
  \otimes\bigotimes_\lambda\mathbf S_{\pi_\lambda}\!\left(\HomG{H_\lambda}{X}\right).
\]
Take duals, tensor with $H_\nu$, and take $G$-invariants on both sides.
Identify $\HomG{H_\lambda}{X}^* \cong \HomG{X}{H_\lambda}$ via composition and use compatibility of Schur functors with duals to obtain
\begin{small}
\[
  \HomG{\Sym^D X}{H_\nu}\cong
  \bigoplus_{(d_\lambda)_\lambda}\bigoplus_{(\pi_\lambda)_\lambda}
  \bigoplus_{(\mu_\lambda)_\lambda}
  \HomG{\bigotimes_\lambda H_{\mu_\lambda}}{H_\nu}
  \otimes\bigotimes_\lambda
    \HomG{\mathbf S_{\pi_\lambda}(H_\lambda)}{H_{\mu_\lambda}}
  \otimes\bigotimes_\lambda\mathbf S_{\pi_\lambda}\!\left(\HomG{X}{H_\lambda}\right),
\]
\end{small}
\!\!\! which is Eq.~\eqref{eq:isotypic-decomposition} after omitting zero summands.
With appropriate normalizations, tracking the elementary tensor
$C\otimes\bigotimes_\lambda L_\lambda\otimes\bigotimes_\lambda T_\lambda$
back through this sequence of isomorphisms gives $Q$ in Eq.~\eqref{eq:isotypic-evaluation}. \qed

\subsection{Proof of Proposition~\ref{prop:isotypic-complexity}}
\label{sec:proof-isotypic-complexity}

Let $h_\lambda=\dim H_\lambda$, let $A_\lambda$ be the largest
edge dimension in $L_\lambda$, and let $R_\lambda$ be the number of terms in $T_\lambda$, as in Appendix \ref{sec:appendix-evaluation}.
If $d_\lambda=0$, let $A_\lambda=R_\lambda=1$. We have the following bounds:
\[
  \begin{array}{lccc}
    & \text{Tableau }T_\lambda & \text{Leaf }L_\lambda & \text{Core }C\\
    \text{Storage}
      & O(d_\lambda)
      & O(d_\lambda2^{h_\lambda}A_\lambda^2)
      & O\!\left(D(\prod_\lambda A_\lambda)^2\right)\\
    \text{Evaluation}
      & O(h_\lambda d_\lambda R_\lambda)
      & O(d_\lambda2^{h_\lambda}A_\lambda^2R_\lambda)
      & O\!\left(D(\prod_\lambda A_\lambda)^2\right)
  \end{array}
\]

The tableau $T_\lambda$ contains $d_\lambda$ entries.
Each of the $R_\lambda$ row-polarization terms requires $O(d_\lambda)$
scalar operations for its coefficient and $O(d_\lambda)$ additions and
scalings of $h_\lambda$-dimensional vectors, hence
$O(h_\lambda d_\lambda R_\lambda)$ flops in total.

The leaf $L_\lambda$ contains $O(d_\lambda)$ tensors.
Each tensor has two dimensions at most $A_\lambda$ and a third at most
$2^{h_\lambda}$. Indeed, the third dimension is
either $h_\lambda$ or a column output dimension, which is bounded by the dimension of an exterior power (see Figure~\ref{fig:isotypic-complexity}(a)).
Each tensor is contracted at most
$R_\lambda$ times (see Appendix~\ref{sec:appendix-evaluation}).

The core $C$ contains at most $D-1$ tensors after omitting
degree zero factors. Each output is an irreducible constituent of the tensor product of its
inputs, so its dimension is at most the product of their dimensions.
Applying this bound along the train, the product of the two input dimensions
at any vertex is at most $\prod_\lambda A_\lambda$. The output dimension
satisfies the same bound, so each tensor has at most
$(\prod_\lambda A_\lambda)^2$ entries
(see Figure~\ref{fig:isotypic-complexity}(b)). Each tensor is contracted once.

\begin{figure}[H]
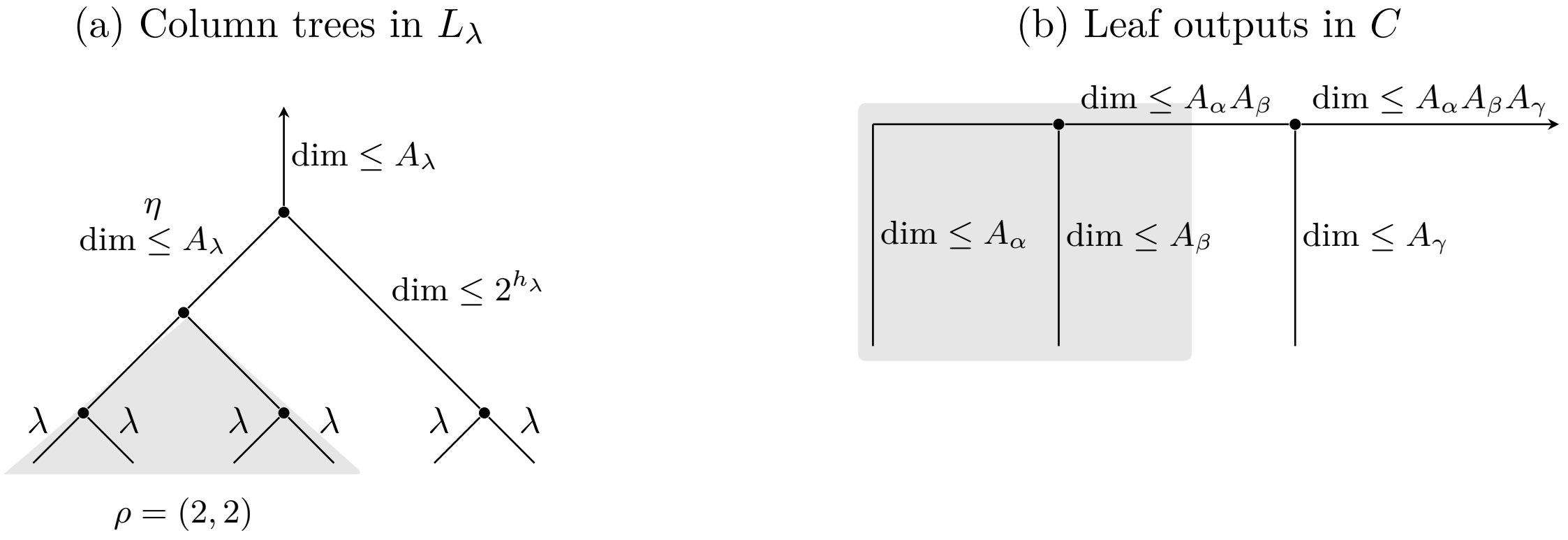

\centering
\tikzset{
  iso proof highlight/.style={iso frame,draw=none,fill=black!10,inner sep=4pt},
  iso proof bound/.style={iso edge label,font=\footnotesize,align=center}
}
\begin{minipage}[t]{.48\linewidth}
\centering
\strut (a) Column trees in $L_\lambda$\par\medskip
\begin{tensordiagram}[iso slot sep=.95,iso output length=1]
  \TensorInput{s1}{0}{}{\lambda}
  \TensorInput{s4}{1}{}{\lambda}
  \TensorInput{s2}{2}{}{\lambda}
  \TensorInput{s5}{3}{}{\lambda}
  \TensorInput{s3}{4}{}{\lambda}
  \TensorInput{s6}{5}{}{\lambda}
  \IsoFuse{column1}{s1}{s4}{}
  \IsoFuse{column2}{s2}{s5}{}
  \IsoFuse{column3}{s3}{s6}{}
  \IsoFuse{partial}{column1}{column2}{}
  \IsoFuse{root}{partial}{column3}{}
  \TensorOutput{}
  \begin{scope}[on background layer]
    \path[iso proof highlight]
      ([xshift=-8pt,yshift=-3pt]tensor-s1) --
      ([xshift=8pt,yshift=-3pt]tensor-s5) --
      ([yshift=5pt]tensor-partial) -- cycle;
  \end{scope}
  \coordinate (proof-rho) at (tensor-partial |- tensor-s1);
  \node[iso metadata,anchor=north] at ([yshift=-8pt]proof-rho)
    {$\rho=(2,2)$};
  \path (tensor-partial) -- (tensor-root)
    node[iso proof bound,auto=left] {$\eta$\\[-1pt]$\dim\le A_\lambda$};
  \path (tensor-column3) -- (tensor-root)
    node[iso proof bound,auto=right] {$\dim\le 2^{h_\lambda}$};
  \path (tensor-root) -- (tensor-output)
    node[iso proof bound,auto=right] {$\dim\le A_\lambda$};
\end{tensordiagram}
\end{minipage}\hfill
\begin{minipage}[t]{.48\linewidth}
\centering
\strut (b) Leaf outputs in $C$\par\medskip
\begin{tensordiagram}[iso slot sep=1.6,iso core sep=2.1,
    iso output length=2.5]
  \TensorInput{a}{0}{}{}
  \TensorInput{b}{1.1}{}{}
  \TensorInput{c}{2.5}{}{}
  \TensorTrain{a,b,c}{{},{}}
  \coordinate (core-cut) at ($(iso-core-2)!.5!(iso-core-3)$);
  \begin{scope}[on background layer]
    \node[iso proof highlight,
      fit=(tensor-a)(tensor-b)(iso-core-1)(iso-core-2)(core-cut)] {};
  \end{scope}
  \path (tensor-a) -- (iso-core-1)
    node[iso proof bound,auto=right] {$\dim\le A_\alpha$};
  \path (tensor-b) -- (iso-core-2)
    node[iso proof bound,auto=right] {$\dim\le A_\beta$};
  \path (tensor-c) -- (iso-core-3)
    node[iso proof bound,auto=right] {$\dim\le A_\gamma$};
  \path (iso-core-2) -- (iso-core-3)
    node[iso proof bound,auto=left] {$\dim\le A_\alpha A_\beta$};
  \path (iso-core-3) -- (iso-output)
    node[iso proof bound,auto=left] {$\dim\le A_\alpha A_\beta A_\gamma$};
\end{tensordiagram}
\end{minipage}
\caption{Dimension bounds for (a) a leaf $L_\lambda$ with
$\pi_\lambda=(3,3)$ (Young symmetrizers omitted) and (b) the core $C$.}
\label{fig:isotypic-complexity}
\end{figure}

For notational simplicity, suppress $\lambda$ where convenient.
To bound $R$, fix a leaf of positive degree and write
$m=\dim\HomG{X}{H_{\lambda}}$ and $M=D+2$.
Let $\alpha_1,\ldots,\alpha_t$ be the multiplicities of the distinct
entries, counted separately in each row. Thus $\sum_j\alpha_j=d$ and
$t\le d$. The tableau has at most $r=\min\{h,m\}$ rows, and row $i$
can use only the entries $i,\ldots,m$. Hence
\[
  t\le b\coloneq\sum_{i=1}^r(m-i+1).
\]
Row polarization gives $\prod_j(\alpha_j+1)$ choices.
Each column has at most $t$ boxes, so its antisymmetrizer uses at most
$t!$ permutations. Applying the AM-GM inequality to $1,\ldots,t$ and to
$\alpha_1+1,\ldots,\alpha_t+1$,
\[
  R\le t!\prod_{j=1}^t(\alpha_j+1)
  \le\left(\frac{t+1}{2}\frac{d+t}{t}\right)^t
  \le(d+1)^t\le M^b,
\]
where the penultimate inequality uses $d\ge t$.

To bound $A$, consider the boxes feeding an edge $H_\eta$. They form
a shape $\rho$ consisting of the first few columns or an initial segment
of one column (Figure~\ref{fig:isotypic-complexity}(a)).
The corresponding subtree restricts to a nonzero map
$\mathbf S_\rho(H_\lambda)\to H_\eta$; otherwise the selected tree
would vanish. This map is surjective because $H_\eta$ is irreducible.
The Schur dimension formula \cite[Theorem~6.3(1)]{FH1991} therefore gives
\[
  \dim H_\eta\le\dim\mathbf S_\rho(H_\lambda)
  =\prod_{1\le i<j\le h}
    \left(1+\frac{\rho_i-\rho_j}{j-i}\right).
\]
Since $\rho$ has at most $r$ rows, only factors with $i\le r$ can
differ from $1$. There are $\sum_{i=1}^r(h-i)=hm-b$ such factors,
each at most $D+1\le M$ since $|\rho|\le D$.
For $h\ge2$, the $h-2$ factors with $i=1$ and $j\ge3$ are at most
$1+D/2=M/2$. Thus
\[
  \begin{aligned}
    A&\le 2^{2-h}M^{hm-b}\le M^{hm-b},\\
    2^hA^2&\le 2^{4-h}M^{2hm-2b}\le4M^{2hm-2b}.
  \end{aligned}
\]
For $h=1$, we have $A=1$ and $hm-b=0$, which also give
$A\le M^{hm-b}$ and $2^hA^2\le4M^{2hm-2b}$.

Since $b\ge1$, these estimates give
\[
  2^hA^2R\le4M^{2hm-b}\le4M^{2hm-1}.
\]
Also $h\le2^hA^2$, so both the tableau and leaf costs in the table
are $O(dM^{2hm-1})$.
Restoring subscripts, $\sum_\lambda d_\lambda=D$ and
$\sum_\lambda h_\lambda m_\lambda=n$ give
\[
  \sum_\lambda d_\lambda M^{2h_\lambda m_\lambda-1}
  \le DM^{2n-1}\le M^{2n}.
\]
For the core, $A_\lambda=1$ on leaves of degree zero and
$A_\lambda\le M^{h_\lambda m_\lambda-1}$ on leaves of positive degree.
There is at least one leaf of positive degree, so
$\prod_\lambda A_\lambda\le M^{n-1}$.
The core and the initial isotypic projections cost $O(DM^{2n-2})$
and $O(n^2)$, respectively. Their combined cost is $O(M^{2n})$. \qed

\end{document}